\documentclass[letterpaper,journal]{IEEEtran}

\usepackage{amsmath,amsfonts}
\usepackage{array}
\usepackage[caption=false,font=normalsize,labelfont=sf,textfont=sf]{subfig}
\usepackage{textcomp}
\usepackage{stfloats}
\usepackage{url}
\usepackage{verbatim}
\usepackage{graphicx}
\usepackage{cite}
\usepackage{booktabs}
\newcolumntype{L}[1]{>{\raggedright\arraybackslash}p{#1}}
\newcolumntype{C}[1]{>{\centering\arraybackslash}p{#1}}
\graphicspath{{figures/}{pictures/}{images/}{./}}
\title{Effects of Locomotion and Viewpoint Control on Immersive Robot Telepresence for Navigation and Inspection}

\author{Ammaar Zaman,
        Neven ElSayed,
        Oliver Prentner,
        Mohamed Ellethy,
        Mostafa M. Lotfy,
        Maram Selsabila Mahmoudi,
        and Eduardo Veas%
\thanks{Ammaar Zaman and Neven ElSayed are with Know-Center GmbH, Graz, Austria. E-mail: azaman@know-center.at; neven.elsaid@gmail.com.}%
\thanks{Oliver Prentner and Eduardo Veas are with Graz University of Technology, Graz, Austria. E-mail: oliver.prentner@tugraz.at; eveas@tugraz.at.}%
\thanks{Mohamed Ellethy is with Siemens Digital Industries Software Egypt. E-mail: mohamed.ellethy@siemens.com.}%
\thanks{Mostafa M. Lotfy and Maram Selsabila Mahmoudi are with the American University in Cairo. E-mail: mostafamlotfy@aucegypt.edu; marammahmoudi@aucegypt.edu.}}

\begin{document}

\maketitle

\begin{abstract}
In many robot telepresence interfaces, movement and viewing are managed through handheld inputs. This can divide attention between steering, looking, and inspection. We compared joystick versus omnidirectional treadmill locomotion and joystick-based versus head-guided camera control in a within-subject \(2 \times 2\) study in which 17 participants used a virtual reality headset to guide a quadruped robot through a physical maze while locating marked wall symbols and capturing them as visual evidence. Mixed-effects analyses showed that head-guided camera control was associated with faster completion, lower wall-contact cost, lower workload, and more favorable presence, usability, technology acceptance, and flow than joystick-based camera control. The strongest overall objective performance was observed in the joystick locomotion plus head-guided camera condition, whereas the omnidirectional treadmill plus head-guided camera condition emerged as the strongest experiential compromise. Exploratory camera analyses showed that head-guided camera control reduced switching, increased reliance on the pan-tilt view, and supported a more stable viewing strategy across the maze. These strategy differences were consistent with part of the performance advantage, especially for wall-contact duration. Targets captured remained near ceiling and simulator sickness remained low across all conditions. Overall, the results indicate that viewpoint control was the stronger differentiator in this combined navigation and inspection task, while locomotion still shaped the objective and experiential tradeoff.
\end{abstract}


\begin{figure*}[!t]
  \centering
  \includegraphics[width=0.9\textwidth]{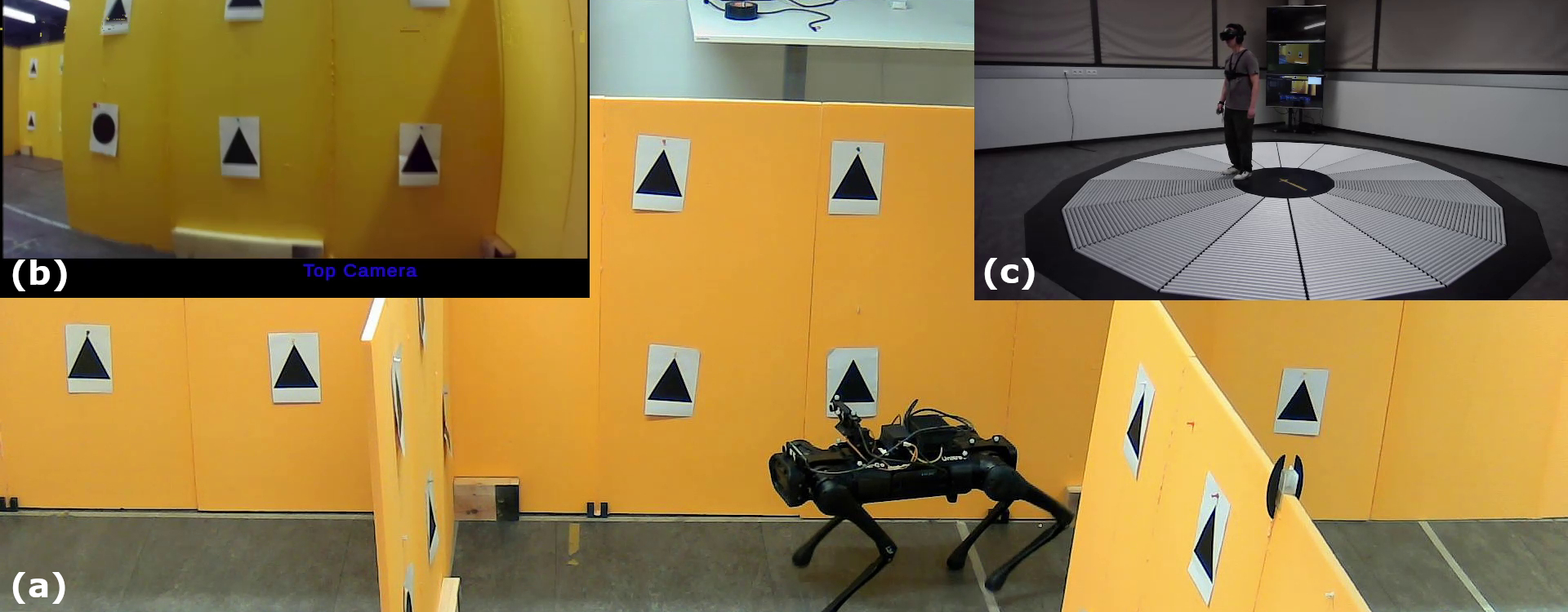}
  \caption{Body coupled locomotion and viewpoint control. (a) A quadruped robot is navigated through a maze by a (c) user walking in an omnidirectional treadmill with head tracking mapped to pan-tilt camera control, capturing (b) a circle target.}
  \label{fig:teaser}
\end{figure*}

\section{Introduction}

Mobile robot telepresence is increasingly relevant in scenarios where people must act in spaces that are distant, hazardous, or difficult to access directly. In these settings, the operator must do more than drive a remote platform from one point to another. The operator must also inspect the environment, notice relevant details, and make decisions through mediated perception. Telepresence has developed from simple mobile video access toward richer robotic systems that support navigation, inspection, collaboration, and remote action across domains such as industry, healthcare, and hazardous field work \cite{Kristoffersson2013A,Youssef2023Telepresence}. In parallel, immersive interfaces have become more prominent in robot operation, with recent reviews showing growing interest in VR based control, visualization, and interaction \cite{Wonsick2020VRReview,Young2019Immersive}. This shift matters because immersive telepresence involves concurrent driving, inspection, and viewpoint management.

In many practical situations, operators must keep making progress while searching for cues, checking objects or structures, and documenting what is found. This creates a dual task burden in which navigation competes with visual inspection for attention and control. Work on multiple resource use and task switching suggests that performance can suffer when perceptual and control demands compete for overlapping resources \cite{Wickens2002MultipleResources,Monsell2003TaskSwitching}. In telepresence, this burden is often expressed through viewpoint management. The operator must decide when to keep moving, when to reorient attention, and how to coordinate robot motion with the currently available view. Prior teleoperation research shows that camera and viewpoint control can directly shape workload, visual attention, usability, and operational performance \cite{Lin2023PerceptionMotion,Liu2025CameraControl,Yoon2025ViewpointControl}.

These challenges drive a shift from telepresence as remote control toward telepresence as body coupled interaction at a distance. Conventional telepresence interfaces control robot motion and viewpoint through handheld input. They can be convenient and familiar, but they divide attention across multiple control loops. A body coupled interface translates body motion to robot control and head motion to viewpoint, linking movement and looking to bodily action. Prior telepresence work connects agency, self-location, and perceptual transparency to how naturally the operator can act through the remote system \cite{Toet2020Embodiment,Nostadt2020Embodiment}. Head-based viewpoint control has shown promise in teleoperated tasks, but the joint role of body coupled locomotion and camera control in immersive robot telepresence is underexplored \cite{Zhu2010HeadOrGaze,Yoon2025Transitions,Yoon2025ViewpointControl}.

We study this question in a robot navigation and inspection task. Participants wear a VR headset and teleoperate a quadruped robot through a physical foam-wall maze, while searching and capturing wall symbols as visual evidence. The study compares conventional joystick control for locomotion and viewpoint with omnidirectional walking and head tracking respectively in a within-subject \(2\times 2\) design. Hereby, whole-body locomotion on an active omnidirectional treadmill is paired with head tracking for loco-viewpoint control of a quadruped robot. Since the robot has no human-like body, the study focuses on the relation between user action, camera control, robot response, and remote visual feedback \cite{Wonsick2020VRReview,Aguilar2025EmbodiedTelepresence}.

Across the mixed-model analyses, head-guided viewpoint control was associated with faster completion, lower workload, and more favorable presence, usability, technology acceptance, and flow than joystick-based viewpoint control. Joystick locomotion with head-guided viewpoint showed the strongest objective performance, while body coupled locomotion with head-guided viewpoint offered the strongest experiential compromise. Target capture was near ceiling and simulator sickness remained low across all conditions. Overall, the results suggest that, in telepresence tasks combining movement and inspection, body coupled locomotion is most useful when the viewpoint control keeps visual inspection easy to maintain.

\section{Background and Related Work}
\label{Related Works}

Robot telepresence combines sensing, communication, and robot control so that an operator can navigate, inspect, manipulate, or collaborate at a distance. Recent reviews describe a shift toward immersive systems that integrate richer sensing pipelines in virtual reality \cite{Kristoffersson2013A,Youssef2023Telepresence}. 
The present study centers on how the operator turns partial remote views into action while the available feedback from the robot continues to change. Situation awareness in remote operation refers to perceiving task relevant elements, understanding their meaning, and anticipating what may happen after the next command.
In a navigation and inspection task, this includes judging clearance, maintaining route understanding, locating targets, and relating each camera view to the robot body.

\subsection{From teleoperation to immersive robot telepresence}

VR has become an increasingly relevant interface for robot operation. A systematic review of VR interfaces for controlling and interacting with robots organizes prior work around visualization, robot control and planning, interaction, and usability \cite{Wonsick2020VRReview}. Across this literature, VR is used to improve 3D understanding, operator performance, situational awareness, and interface intuitiveness during remote tasks \cite{Wonsick2020VRReview,stotko2019vr}. Prior immersive systems also proved valuable for remote exploration, mobile robot operation, and telemanipulation, especially with spatially consistent views and control coupled to the user's body motion \cite{lipton2017baxter,kuo2021slam,schwenk2025immersive}. For navigation-oriented telepresence, the relevant question is how naturally the operator can move, look, and act through the remote platform.

\subsection{Camera control and viewpoint management}
A central issue in this literature is that camera and viewpoint control are not secondary interface details. They shape how operators divide attention between moving, searching, and acting. In many robot tasks, the operator must balance progress while deciding where to look next and interpreting the relation between the view and the robot's motion. This makes viewpoint management part of the task itself rather than a passive support layer \cite{almeida2020interface,Lin2023PerceptionMotion}.

This issue becomes especially important in navigation-and-inspection tasks, where users must coordinate locomotion with ongoing visual selection. Viewpoint control can itself become a source of control burden because operators may need to decide whether to move the robot, reorient the camera, switch views, or pause to recover spatial understanding. Earlier comparisons of camera and motion control in teleoperated robots show that the control mapping assignment affects task performance and operator workload \cite{doisy2017comparison}. Recent work links active camera control to perception-motion coupling strategies, cognitive load, visual attention, usability, and operational performance \cite{Lin2023PerceptionMotion,Liu2025CameraControl}. Studies on head movement, gaze, and other viewpoint control techniques show that remote camera control can change both performance and subjective preference \cite{Zhu2010HeadOrGaze,Yoon2025ViewpointControl}. Work on viewpoint transitions further suggests that deciding when to move between views or control modes is itself a meaningful part of teleoperation behavior \cite{Yoon2025Transitions}.

Taken together, these studies suggest that camera control is both a visualization choice and part of the sensorimotor coupling between the operator, the interface, and the remote robot. This is directly relevant to the present work, where one of the key questions is whether a more body coupled viewpoint mapping can reduce the burden of managing visual perspective during robot navigation and inspection.

\subsection{Spatial understanding and body coupled telepresence}

Operators often have to infer layout, orientation, and future movement consequences through a limited set of remote views. Spatial understanding becomes fragile when visual and motor reference frames do not align. 
Prior work showed that remote virtual environments can differ from real environments in spatial presence, performance, and behavior \cite{khenak2020spatial}. 
Other teleoperation work notes that restricted viewpoints, frame-of-reference conflicts, and limited sensory context reduce spatial awareness and increase control burden \cite{almeida2020interface,mutzenich2021updating,fabris2021immersive}.

This problem is not limited to what is visible in the current frame. Spatial knowledge also involves maintaining route structure, object locations, heading, and relations among the user, the viewpoint, the robot, and the surrounding scene. Cognitive map research describes how spatial knowledge supports navigation, memory, and flexible future behavior \cite{epstein2017cognitive,peer2020structuring,behrens2018cognitive}. In robot telepresence, spatial knowledge has to be built from mediated camera views and updated while the robot moves.

For navigation and inspection, these issues are especially important because, while users move and interpret the scene, should they decide to reorient they must map the control to the camera or robot. Here the idea of body coupled telepresence becomes relevant. Recent work argues that performance improves when the operator experiences stronger agency, self-location, and body-referenced coupling with the remote system \cite{Toet2020Embodiment}. Broader telepresence and embodied interaction literature discusses the intersection between  presence and bodily experience in mediated systems and why body coupled control mappings matter for remote interaction \cite{Nostadt2020Embodiment,gonzalezfranco2018avatar}.

This perspective is not limited to strictly humanlike remote bodies. Research on agency and body-referenced coupling suggests that these experiences can extend beyond highly anthropomorphic forms, which makes this line of work relevant for humanoid teleoperation and for telepresence through mobile legged robots. Prior work on perspective and body related experience in VR demonstrated that viewpoint, agency, and self-location can separate from one another. First-person views support strong body related experience and accurate interaction, while third-person or alternating viewpoints may support broader spatial awareness \cite{gorisse2017first,debarba2017characterizing}. This distinction is relevant for this study where the experimental question concerns whether useful action feedback and viewpoint coupling can still support remote operation if the robot is not humanoid.

Hence, the present system is termed partial body coupled telepresence. The operator's head motion maps to a mechanical pan-tilt camera, and body motion on the treadmill maps to bounded quadruped velocity commands. The study asks whether these mappings provide useful action feedback from a non-human robot body during navigation and inspection.

\subsection{Locomotion and omnidirectional treadmills}

A second major design axis in immersive telepresence concerns how the user moves. Controller-based locomotion is common due to its simplicity and convenience, but it does not provide the same bodily cues as walking. VR locomotion research has shown that locomotion technique affects presence, cybersickness, spatial knowledge, and navigation behavior \cite{boletsis2019vr,langbehn2018evaluation,buttussi2021locomotion}. A broader typology of VR locomotion techniques emphasizes differences in movement continuity, body involvement, and control mapping \cite{Boletsis2022Typology}. Recent work further shows that locomotion choice influences spatial orientation and user experience during navigation tasks \cite{kim2024locomotion,ho2025virtual}.

This distinction is useful because travel is only one part of the wider navigation task. A locomotion technique may move the user efficiently, but the same technique must also preserve route continuity, support inspection, and leave enough attention for the next action. This is why locomotion cannot be interpreted only as a scale from less bodily to more bodily input. Its value depends on whether movement, visual feedback, and task demands fit together.

Omnidirectional treadmill systems differ from one another. Active omnidirectional treadmills such as Omnifinity Omnideck move the walking surface keeping the user near the center and tracking body motion for input. Passive omnidirectional treadmills rely on low friction surfaces and user generated foot motion. Belt based systems such as CyberWalk use controlled belt motion to support walking within a mechanical platform \cite{souman2008cyberwalk}. They all differ from unconstrained real walking because the device constrains gait, balance, speed, comfort, adaptation, and who can use the system safely \cite{lewis2024gait,Chakraborty2024ODT}.

Within this literature, omnidirectional treadmills are interesting because they reintroduce walking-like input while keeping the user within a constrained physical space. Prior work has examined their usability, cybersickness, gait characteristics, and navigation performance in virtual environments \cite{hooks2020evaluating,lohman2022evaluating,Chakraborty2024ODT}, including gait and movement behavior during treadmill-based virtual walking \cite{lewis2024gait}. Most of the past research focused on virtual navigation rather than immersive robot telepresence. It remains unclear whether the benefits often associated with more body coupled locomotion carry over directly to robot telepresence tasks that combine movement with inspection.

\subsection{Gap and motivation}

Prior work shows that viewpoint control affects teleoperation performance and experience, while locomotion technique modulates behavior in immersive environments \cite{Youssef2023Telepresence,Wonsick2020VRReview}. These lines of work are often studied separately. What remains less explored is how walking-based locomotion and head-guided camera control jointly affect a telepresence task that requires both movement and inspection. This gap is especially relevant for non-humanoid remote bodies such as quadruped robots, where body coupling, locomotion, and viewpoint control do not map cleanly onto a human avatar.

\section{Research Focus and Contributions}
When exploring an environment, people naturally use head motion and locomotion together while acquiring information. This paper studies such phenomenon in an immersive robot telepresence task. The study is guided by three expectations from the related work: head guided viewpoint control may reduce camera-management burden during inspection, body coupled locomotion may support presence and flow while adding physical effort and adaptation demands, and the interface that feels best may not be the interface that produces the fastest or safest task performance.

\textbf{\textit{RQ1}} In a telepresence task that combines navigation and inspection, how do body coupled and manual locomotion, combined with active or manual viewpoint control, affect task efficiency, wall contact safety, and target capture?

\textbf{\textit{RQ2}} How do locomotion and viewpoint control shape workload, presence, simulator sickness, usability, technology acceptance, and flow during immersive robot telepresence?

\textbf{\textit{RQ3}} When operators can use a robot body aligned view and an independently steerable view, what camera use strategies emerge, and how do those strategies relate to performance and experience?

The study contributes a controlled comparison of techniques for locomotion and viewpoint control in immersive robot telepresence, empirical findings on performance and user experience, and behavioral evidence relating camera-use strategies relate to those outcomes. These contributions frame immersive telepresence as a coordination problem in which movement, viewpoint control, and remote action must work together.

\section{Experimental Setup}
\subsection{Robot platform}
The experiment used a Unitree AlienGo quadruped robot. It has two onboard computers for low-level and high-level control. High-level motion control is handled by the onboard Mini PC, which has access to robot state information and implements the control interface used in this study, together with directional movement commands. The robot was used in \textit{walking} mode for active locomotion and received bounded target velocity commands f(forward: fwd, backwards: bwd, strafe left: lstr, strafe right: rstr, rotate left: lrot, rotate right: rrot). The command limits were 0.5~m/s for forward, backward, and lateral motion, and 1.5~rad/s for yaw. Robot control is realized over WiFi through the high-level control computer.

\begin{figure}[!t]
  \centering
  \includegraphics[width=\linewidth]{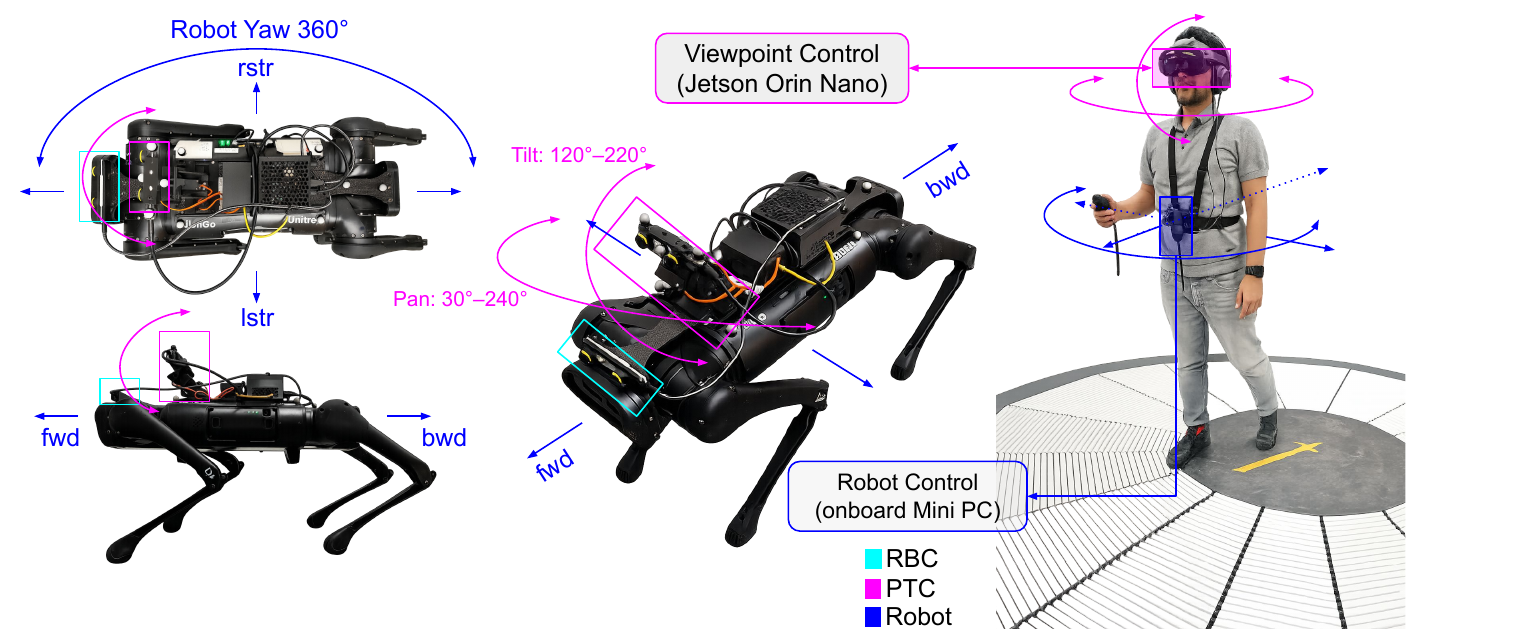}
  \caption{Overview of the robot platform and telepresence interface.}
  \label{fig:Robot}
\end{figure}

For robot safety, the controller monitors IMU roll and pitch, gyroscope and accelerometer signals, foot-force unloading, and passive-drop conditions. On a safety trigger, the robot enters a safety routine with zero command velocity and transitions to damping mode if stability does not resolve.

The robot platform is extended with two stereo cameras (ZED Mini) offering two views: a lower fixed Robot Body Camera (RBC) mounted at the head of the robot and an independently controllable Pan-Tilt Camera (PTC) mounted above the torso on a pan-tilt system actuated by two MG996R servomotors mounted in a 3D printed frame. The cameras connect via USB 3.0 to an Nvidia Jetson Orin Nano viewpoint controller mounted on the robot torso and powered from the robot core. The pan-tilt system applied soft servo limits of 30--240$^\circ$ for pan and 120--220$^\circ$ for tilt. The two cameras are calibrated using a standard checkerboard-based stereo calibration procedure so that their forward views align to a common straight-ahead reference when centered. The PTC is positioned so that looking downward retains visibility of the robot body width. The camera heights and quadruped body geometry did not reproduce a human eye height or human body shape, so the apparatus provided partial body coupled telepresence rather than a humanoid telepresence setup. The viewpoint controller communicates through WiFi, receiving commands, updating pan-tilt status, and streaming video from both synchronized stereo cameras. The video pipeline operates at 15~Hz, capturing stereo frames at 2560 \(\times\) 720, resizing them to 960 \(\times\) 540, and compressing them with JPEG quality 70 for transmission. At any moment the system transmitted the active camera stream used by the operator rather than treating both camera views as separate experimental video channels.

\subsection{Telepresence platform}\label{sec:telepresence_platform}
The telepresence application defines the interface for capturing the operator input for motion and viewpoint control and transferring them to the robot while rendering the robot viewpoints for the operator. It is implemented in Unity3D. The application receives streams from each left and right unit in the stereo camera separately and renders them on the corresponding left and right eye textures for stereo display in the headset. The system offers two locomotion modalities (joystick, walking) and two modalities for viewpoint control (joystick, head). Velocity and camera-control commands are emitted at 20~Hz.
The four interface combinations are abbreviated as JJ for joystick locomotion with joystick camera control, JH for joystick locomotion with head guided camera control, OJ for Omnideck locomotion with joystick camera control, and OH for Omnideck locomotion with head guided camera control.

\noindent\textbf{Locomotion modalities.}
The telepresence application generates velocity commands according to the locomotion modality. Two modalities are used. In joystick locomotion, controller axes are converted into target robot velocities. In treadmill locomotion, tracked body motion on the omnidirectional treadmill is converted into target robot velocities. Instead of a direct distance mapping between operator and robot travel, velocity commands are used because they match the robot high level walking interface and allow both locomotion inputs to share the same robot command path. In both modalities the Unity application smoothed small changes, suppressed jitter near zero input, and constrained commands before sending them to the robot. The robot side used the most recent received command rather than executing a queued sequence of motions.

\textit{Locomotion through joystick}. VR controllers control forward, lateral, and yaw motion directly. The left thumbstick controls forward motion and yaw, the right thumbstick controls lateral motion. Hereby, joystick and treadmill conditions share the same robot command type after input processing, making locomotion comparison about the human input mapping rather than a different robot walking controller.

\textit{Locomotion through omnidirectional treadmill}, robot motion is derived from tracked body translation and yaw on the treadmill. The left controller is attached to the torso with an action-camera body strap to provide the body reference direction.Treadmill motion provides the basis for computing robot velocity updates. The mapping therefore used the participant's body motion as a continuous velocity input while the treadmill kept the participant near the center of the local walking area.

\noindent\textbf{Viewpoint control modalities.}
Viewpoint control is tied to robot control for the RBC. Its viewpoint follows robot position and body orientation. The right controller primary button switches between RBC and PTC. The latter is controlled by:

\textit{Joystick viewpoint control} uses the right joystick to control pan (left/right) and tilt (up/down). Small joystick inputs were ignored by a dead zone, and valid inputs updated the pan and tilt target within the servo limits.

\textit{Head viewpoint control} uses head orientation tracking from the headset as control commands for the PTC. Pan and tilt are derived from headset orientation relative to a reference frame using a 1.5$^\circ$ dead zone and a recenter function. The pan and tilt controller smoothed and clamped updates to the same physical range used in the joystick camera condition. Recenter commands returned the PTC to its forward reference.

\noindent\textbf{Mixed locomotion--viewpoint control modalities.}
In the JJ condition, joystick controls for locomotion conflict with joystick controls for viewpoint. A modal switch (left-controller Y button) toggles between robot-control and pan-tilt control modes. No such conflict exists in other conditions. In OJ, locomotion is controlled independently through treadmill motion. In  JH, pan and tilt follow head motion independent from joystick locomotion. In OH, treadmill locomotion is independent of pan--tilt control through head motion. The RBC remains body aligned in all conditions. Robot motion was body relative in all conditions and independent from the selected camera view. Participants could move the robot while viewing either the RBC or the PTC. Thus, JJ represents a fully manual baseline with shared controller resources. Other manual mappings with separate camera controls could reduce the cost of switching between robot and camera control.

\begin{table*}[t]
\centering
\footnotesize
\setlength{\tabcolsep}{3pt}
\caption{Interface mappings and command limits used in the four experimental conditions. The table separates the operator input, command behavior, and condition role for locomotion, viewing, active viewpoint control, camera switching, and recentering.}
\label{tab:interface-mapping}

\begin{tabular}{@{}p{0.18\textwidth}p{0.24\textwidth}p{0.32\textwidth}p{0.20\textwidth}@{}}
\toprule
Component & Operator input & Command behavior & Condition role \\
\midrule
Joystick locomotion & Left thumbstick for forward motion and yaw. Right thumbstick for lateral motion. & Controller axes were mapped to bounded target velocity commands. Command limits were 0.5~m/s for forward, backward, and lateral motion, and 1.5~rad/s for yaw. & Used in JJ and JH. \\
Omnideck locomotion & Tracked torso translation and yaw while standing on the Omnifinity Omnideck 5. & Body motion was mapped to target velocity commands rather than one to one distance travel. Commands were smoothed, thresholded near zero input, and saturated at the same velocity limits as joystick locomotion. & Used in OJ and OH. \\
Robot Body Camera & Camera switch selected the fixed lower view. & The RBC was a view, not an active camera-control mode. It was aligned with the robot body and supported heading, wall clearance, and recovery of robot body orientation. & Available in all conditions. Robot motion remained body relative. \\
Joystick PTC control & Right thumbstick pan and tilt input when PTC control was active. & Joystick input used a dead zone and updated bounded pan and tilt targets within the 30--240$^\circ$ pan and 120--220$^\circ$ tilt soft servo limits. & Used in JJ and OJ. In JJ active robot control and active PTC control were serial because of the mode switch. \\
Head guided PTC control & Headset yaw and pitch relative to a reference frame. & Head motion was mapped to bounded pan and tilt targets within the same soft servo limits. Fast head turns could saturate or lag within the configured PTC range rather than producing unbounded camera motion. & Used in JH and OH. Locomotion and active PTC manipulation could occur together. \\
Camera switching and recentering & Right controller button for view switching and controller recenter commands. & The active view switched between RBC and PTC. Recenter commands returned the PTC to its forward reference. & Available in all conditions. \\
\bottomrule
\end{tabular}

\end{table*}

\subsection{Communication Architecture and Data Flow}\label{sec:communication}

The telepresence system is implemented as a distributed local-network architecture centered on a Dockerized Private Cloud (DPC) running on a dedicated local workstation. The DPC is the main on-premise orchestration layer for communication between distributed robotic components \cite{ellethy2023digital}. It decouples telepresence controls, transmitted through persistent Socket.IO connections to the DPC, from communication with ROS-based robot services. This allows Unity-based components and robot-side processes to exchange commands and telemetry within the same communication fabric.

\subsection{Robot Navigation and Inspection Task}\label{sec:task}
A robot navigation and inspection task is implemented to study the effects of two locomotion modalities and two viewpoint-control modalities under a \(2\times 2\) design. Participants teleoperate a Unitree AlienGo quadruped robot through a physical foam-wall maze while wearing a VR headset. The task goal is to move the robot from start point to end point while capturing wall symbols as visual evidence. The task combines route completion, viewpoint control, visual search, and target documentation within the same trial.

The maze has a fixed outer footprint of 4.2\,m \(\times\) 9.0\,m and is built from 1.25\,m high foam walls. Wall segments are modular, and the study uses 1.2\,m and 1.8\,m wide wall sections. Openings for entry, exit, and turns are 1.2\,m wide, all turns are 90 deg. Three maze variants are created by shifting the position of openings relative to the start and end direction while preserving the same overall footprint, so that participants cannot rely on memorizing a single repeated layout.

Targets are placed across the maze walls according to a predefined schedule. Circle symbols serve as task-relevant targets to be captured in screenshots, triangle symbols populate the remaining wall locations as non-target distractors. A trial contains 5, 6, or 7 circle targets, balanced across the 12 trials completed by each participant. The traversable maze is divided into five ordered sections from start to end, hereby the target schedule balances upper-versus-lower placement across these sections. This varies target visibility across viewpoints and robot positions without favoring any one interface condition.

Two camera views are available: the fixed Robot Body Camera (RBC) and the independent Pan-Tilt Camera (PTC). Participants are free to switch and use either. The two views differ in geometric affordances. Being coupled to robot heading, what enters the RBC view depends on the robot body pose relative to the maze wall. It hereby offers a body aligned view for steering, wall clearance, and recovering the robot heading. But some targets may require additional body reorientation or larger viewing distance to become fully visible. Instead, the PTC allows participants to redirect the viewpoint without changing robot heading in the same way. The PTC is served an independent inspection view for search and target capture while the robot could remain oriented for navigation. Hence, it reduces the need to rotate or reposition the robot body simply to bring a target into view, making viewpoint management part of the task itself rather than a separate support function. The task is a controlled navigation and inspection scenario focused on movement, active viewpoint management, and visual documentation in robot telepresence \cite{Youssef2023Telepresence,Wonsick2020VRReview}.

\noindent\textbf{Calibration, Safety, and Failure Handling.}
The task was implemented as an extension of the telepresence platform, including a virtual replica of the maze and robot in Unity, aligned to the physical environment through Vicon-based tracking. This allowed physical robot motion and maze contacts to be mirrored in the virtual scene and logged as collision events. Participants navigated from the live stereo robot camera feed rendered in the headset rather than from the simulated maze. The Unity replica supported alignment, triggers, logging, contact detection, experiment monitoring, and safety checks. Between trials, the physical maze was manually re-aligned to predefined floor markings and the robot was returned to the start position and orientation. Before each trial, participants were asked to verify that they could switch cameras, move the robot using the locomotion modality of the current condition, and capture a sample screenshot. 

During the study, a second experimenter monitored robot safety and intervened when wall contact posed physical risk to the robot. Otherwise, wall contacts were tolerated, logged, and the foam maze was re-aligned after each trial. Participant safety on the treadmill relied on a spotter and an emergency push-stop button at the operator station. No harness was used. 

\noindent\textbf{Latency and Synchronization Logging.}
The system logged command and stream telemetry at multiple stages of the control pipeline. Before each block, a time-health script verified operating-system time services across all computers. Latency reporting was based on application-layer round-trip time (RTT). The latency values therefore characterize command acknowledgment and local receive to apply timing, not the full end to end visual delay experienced by the participant.

Latency was summarized within each trial and then aggregated across logs for each channel. Table~\ref{tab:latency-summary} reports the median of trial medians, the IQR of trial medians, and the median trial \(p95\). Robot receive-to-apply values were derived from matched robot-side logs. The other channels derived from Unity-side trial logs. The video stream operated at 15Hz. Glass to glass latency was not measured, end to end visual delay was not inferred from RTT values. Condition-wise latency differences were small enough to consider it stable across conditions.

\noindent\textbf{Apparatus.}
Unity3D (v2021.3.15f1) was used to implement the task. It ran on a Meta Quest Pro headset with standard controllers over a Steam Link connection. Rendering used 4128 \(\times\) 2096 resolution at 72~Hz. Interpupillary distance was adjusted before each session using the Quest fit procedure. Participants stood on the base platform of an Omnifinity Omnideck 5 active omnidirectional treadmill in all conditions and wore noise-cancelling headphones with monotonous white noise. The setup was tracked in Vicon Tracker.

\begin{table}[t]
\centering
\footnotesize
\setlength{\tabcolsep}{2pt}
\caption{Trial-level latency characterization by logged channel.}
\label{tab:latency-summary}
\begin{tabular}{@{}p{0.36\columnwidth}C{0.18\columnwidth}C{0.22\columnwidth}C{0.18\columnwidth}@{}}
\toprule
Channel & Median (ms) & IQR (ms) & Trial \(p95\) (ms) \\
\midrule
Robot command RTT & 9.02 & 8.82--9.55 & 35.95 \\
Robot receive-to-apply & 1.02 & 0.99--1.04 & 1.91 \\
Pan-tilt RTT & 15.22 & 15.00--15.42 & 21.63 \\
Pan-tilt receive-to-apply & 3.00 & 2.00--3.00 & 6.00 \\
Camera sender ACK RTT & 19.53 & 18.98--21.00 & 30.70 \\
Camera receive-to-apply & 7.90 & 7.52--8.09 & 13.95 \\
\bottomrule
\end{tabular}
\end{table}

\section{Experimental Design and Procedure}\label{sec:procedure}

\subsection{Participants and Ethical Considerations}
A total of 17 participants (12 men, 5 women) aged  $M=29.59$ years ($SD = 6.05$, $range = 20-48$) completed the study. All of them had normal or corrected-to-normal vision. Eleven participants reported prior VR experience, five reported experience operating a robot or remote-controlled device, and six reported prior exposure to omnidirectional treadmills. They all completed all 12 trials, for a total of 204 complete trials.

Participants were recruited through campus flyers, mailing lists, and student-group recruitment channels. Inclusion criteria were age 18 years or older, normal or corrected-to-normal vision. Exclusion criteria were pregnancy, self-reported susceptibility to motion sickness or VR discomfort, current use of medication likely to impair balance, alertness, or comfort during the study, or any self-reported condition that would make treadmill use unsafe. All participants received voucher compensation for their participation. All participants provided informed consent before the study. The study was approved by the institutional Ethics Committee.

\subsection{Repeated-Measures Design}\label{sec:design}
The study used a \(2\times 2\) within-subjects design with locomotion modality (Joystick vs Omni) and viewpoint (camera) control (Joystick vs Head) as independent variables. This produced the four interface combinations defined in Section~\ref{sec:telepresence_platform}, JJ, JH, OJ, and OH. The conditions represent a systematic analysis of body coupling with different bodily links in viewpoint and locomotion \cite{Aguilar2025EmbodiedTelepresence}.

Each participant completed four condition blocks, one per interface condition. Block order refers to the serial position of a condition block within the full session (1--4), counterbalanced across participants using a Latin-square schedule. Within each block, participants completed three trials, one in each maze variant; trial in block refers to the within-condition trial position (1--3). Maze refers to the three maze variants, and target count refers to whether a given trial required 5, 6, or 7 targets to be captured. Each participant therefore completed 12 trials in total. Maze variant and target count were part of the task schedule rather than primary independent variables. They were included to vary route geometry and inspection demand while preserving the main \(2\times2\) interface comparison. A participant-specific counterbalancing schedule balanced condition order and maze order across the sample, while a predefined target-assignment schedule balanced inspection demand and upper-versus-lower target placement without favoring any one interface condition. In the final dataset, each interface condition contributed 51 trials, each maze variant appeared 68 times, and trials requiring 5, 6, or 7 targets each appeared 68 times.

\subsection{Training and Familiarization}
Before each condition block, participants completed a short training session designed to match the upcoming interface condition. It consisted of five structured scenarios that required basic robot movement, camera switching, screenshot capture, and navigation while using the full range of motions supported by the current locomotion and viewpoint mappings. The scenarios included moving between predefined cardinal positions, switching between cameras, capturing sample screenshots, and navigating while capturing screenshots. Participants were also instructed to account for the robot body footprint while navigating. Training was criterion based. It typically lasted approximately 10 minutes per block and continued until the participant successfully completed at least four of the five scenarios. No participant required training beyond this criterion. This familiarization reduced unsafe or confused use, but it did not remove practice effects or the possibility that the four interfaces had different adaptation demands. 

\subsection{Experimental Procedure}
Participants stood at the center plate of the omnidirectional treadmill in all conditions. Using the same standing posture across conditions kept the operator station, headset use, camera views, and safety procedure consistent. They wore the VR headset and noise-cancelling headphones with white noise during task execution. Control mappings and camera-operation behavior followed the interface configuration described in Section~\ref{sec:telepresence_platform}. Each trial began at the marked maze start point and ended when the robot reached the end point. Participants were instructed to complete the maze as fast as possible, avoid colliding with the walls, and capture all required wall targets within their intended maze sections using one target per screenshot. There was no time limit per trial. Because both views could support both navigation and target capture, participants were not instructed to assign different task roles to the two cameras. Instead, they were free to use the available viewpoints in whatever way they found most effective for completing the route and documenting the targets. Scheduled breaks were provided between trials and condition blocks, typically lasting 5--10 minutes. Each condition block lasted approximately 50--55 minutes, and the full session lasted approximately 3.5 hours.

\subsection{Screenshot Scoring and Questionnaires}
After each trial, participants removed the headset and completed the post-trial questionnaire battery on a tablet. The questionnaire order was equipment comfort and technical issues, NASA-TLX items, simulator sickness symptom ratings, PQ items, SUS items, TAM items, Flow items, and task-specific camera/control items. The sickness items were administered after each trial only and averaged into a symptom score on the original 0 to 3 item scale.

Screenshot files were stored with filenames linked to trial metadata and the active camera. After the experiment, screenshot completion was derived against the predefined target schedule using a semi-automated workflow based on the stored filenames, the participant-specific target-assignment tables, and documented adjustments when a single screenshot contained more than one valid target. Triangle distractors were excluded from the target count. Repeated captures of the same target were counted once for completion. When one screenshot contained more than one scheduled circle target, each visible scheduled target was credited after manual validation. Screenshots that could not be matched to the target schedule were reviewed against the trial metadata and excluded from completion scoring when they showed distractors or unusable captures. In the analysis reported here, target-completion percentages were capped at the required target total for estimation. The resulting target capture measure is a trial level completion score for visual documentation, especially because performance was near ceiling.

\begin{table*}[t]
\centering
\footnotesize
\setlength{\tabcolsep}{4pt}
\caption{Post-trial questionnaire measures and scoring.}
\label{tab:questionnaire-scoring}

\begin{tabular}{@{}p{0.22\textwidth}p{0.16\textwidth}p{0.18\textwidth}p{0.34\textwidth}@{}}
\toprule
Measure & Items & Response range & Reported score \\
\midrule
NASA-TLX workload & 6 & 1 to 10 & Raw item mean. Lower values indicate lower workload. \\
Simulator sickness symptom score & 11 symptom ratings & 0 to 3 & Mean across symptom ratings. Lower values indicate fewer symptoms. \\
Presence questionnaire & 6 & 1 to 10 & Item mean. Higher values indicate higher presence. \\
System usability scale & 10 & 1 to 5 & Standard SUS conversion to a 0 to 100 score. Higher values indicate higher usability. \\
Technology acceptance & 5 & 1 to 10 & Item mean. Higher values indicate higher acceptance. \\
Flow & 5 & 1 to 10 & Item mean. Higher values indicate higher flow. \\
\bottomrule
\end{tabular}

\end{table*}

\subsection{Measures}\label{sec:measures}
\noindent\textbf{Independent Variables}
The two primary within-subject independent variables were locomotion (omni vs joystick) and viewpoint modality (head vs joystick). The mixed-model analyses included maze variant (Maze 1--3), target count (5, 6, or 7 required targets), block order (condition-block position 1--4), and trial in block (within-block trial position 1--3) as additional fixed effects to account for route variation, inspection demand, and learning effects.

\noindent\textbf{Objective Performance Measures.}
The primary objective performance outcomes were chosen to reflect efficiency, safety, and inspection success in the telepresence task. Completion time was defined as elapsed time from the maze start trigger to the maze end trigger. Targets captured was defined as trial level screenshot completion relative to the required target count for that trial, using the screenshot scoring described above. Safety was quantified using wall-contact count and wall-contact duration derived from the logged collision events. Together, these measures capture whether participants could complete the route efficiently, avoid wall contact, and document the required targets while navigating.

\noindent\textbf{Subjective Measures.}
Subjective experience was assessed after each trial using post trial questionnaire scales for workload, simulator sickness symptoms, presence, system usability, technology acceptance, and flow. Lower values indicate more favorable workload and symptom outcomes, whereas higher values indicate more favorable presence, usability, technology acceptance, and flow outcomes. The item counts, response ranges, and scoring rules are summarized in Table~\ref{tab:questionnaire-scoring}. Exploratory task-specific items captured perceived coordination, intuitiveness, comfort, and camera reliance, and a post-study questionnaire collected overall difficulty ratings, preference rankings, and feedback.

\noindent\textbf{Camera-Behavior Measures.}
The system also recorded camera-behavior measures to characterize how participants managed the RBC and PTC. The main trial-level camera outcomes analyzed in this paper were camera switching frequency, camera dwell allocation, capture camera share, use of both cameras, and camera viewing time. Camera switching, dwell allocation, use of both cameras, and camera-viewing-time measures were computed during active maze traversal between the start and end triggers. Capture camera share was computed from the target-capture screenshots saved for a trial, including valid pre-start captures. Each maze was divided into five ordered sections from start to end so that camera dwell allocation and camera switching frequency could also be examined across early, middle, and late parts of the route.

\subsection{Statistical Analysis}\label{sec:statistical-analysis}
The analyses used trial level mixed models because each participant contributed repeated observations across conditions, maze variants, and trial positions. Participant was included as a random intercept in all primary models. Locomotion and camera control were the primary fixed effects. Maze variant, target count, block order, and trial in block were included as task and order controls because they reflected route variation, inspection demand, and practice within the session. Gaussian outcomes were analyzed with linear mixed models. Positive continuous, count, bounded completion, and zero heavy outcomes were analyzed with generalized linear mixed models using Gamma, negative binomial, grouped binomial, or Tweedie families as appropriate. Model comparisons were used to decide whether to retain the locomotion by camera control interaction. Type III tests summarized fixed effects, and estimated marginal means on the response scale were used for condition summaries and Bonferroni adjusted pairwise comparisons.

\section{Results}
Results are reported for the objective performance measures, questionnaire outcomes, and exploratory camera behavior measures. The abbreviations JJ, JH, OJ, and OH refer to the four interface conditions defined in Section~\ref{sec:telepresence_platform}. Omnibus model results are summarized in Table~\ref{tab:model-summary}, and model based estimated marginal means on the response scale are summarized in Table~\ref{tab:objective-descriptives}.

\begin{table*}[t]
\centering
\scriptsize
\setlength{\tabcolsep}{2.5pt}
\renewcommand{\arraystretch}{1.08}
\caption{Omnibus mixed model summary for the main outcomes. Significant fixed effects are shown with Type III Wald tests. \(R^2_m\) and \(R^2_c\) denote marginal and conditional \(R^2\).}
\label{tab:model-summary}

\begin{tabular}{@{}L{0.13\textwidth}L{0.17\textwidth}L{0.28\textwidth}L{0.28\textwidth}L{0.09\textwidth}@{}}
\toprule
Outcome & Model & Interface effects & Task and order effects & \(R^2_m\), \(R^2_c\) \\
\midrule
Completion time & Gamma GLMM, log &
Locomotion, \(\chi^2(1)=21.51\), \(p<.001\) \newline
Camera control, \(\chi^2(1)=119.96\), \(p<.001\) &
Block order, \(\chi^2(3)=34.26\), \(p<.001\) \newline
Trial in block, \(\chi^2(2)=16.91\), \(p<.001\) & .304, .696 \\
Target completion & Grouped binomial GLMM, logit &
No fixed effect reached significance \newline
Locomotion, \(\chi^2(1)=3.44\), \(p=.064\) \newline
Camera control, \(\chi^2(1)=1.14\), \(p=.285\) &
Maze, \(\chi^2(2)=0.26\), \(p=.878\) \newline
Target count, \(\chi^2(2)=2.84\), \(p=.241\) \newline
Block order, \(\chi^2(3)=6.94\), \(p=.074\) \newline
Trial in block, \(\chi^2(2)=2.81\), \(p=.246\) & .477, .830 \\
Wall contact count & Negative binomial GLMM, log &
Locomotion, \(\chi^2(1)=3.87\), \(p=.049\) \newline
Camera control, \(\chi^2(1)=14.95\), \(p<.001\) \newline
Locomotion by camera control, \(\chi^2(1)=12.64\), \(p<.001\) &
Block order, \(\chi^2(3)=8.23\), \(p=.041\) & .161, .348 \\
Wall contact duration & Tweedie GLMM, log &
Camera control, \(\chi^2(1)=16.92\), \(p<.001\) &
Block order, \(\chi^2(3)=14.00\), \(p=.003\) & .171, .473 \\
NASA TLX & Gaussian LMM &
Locomotion, \(\chi^2(1)=16.73\), \(p<.001\) \newline
Camera control, \(\chi^2(1)=32.00\), \(p<.001\) &
Block order, \(\chi^2(3)=21.46\), \(p<.001\) \newline
Trial in block, \(\chi^2(2)=7.58\), \(p=.023\) & .091, .767 \\
Presence & Gaussian LMM &
Locomotion, \(\chi^2(1)=23.19\), \(p<.001\) \newline
Camera control, \(\chi^2(1)=68.93\), \(p<.001\) \newline
Locomotion by camera control, \(\chi^2(1)=13.14\), \(p<.001\) &
Block order, \(\chi^2(3)=29.91\), \(p<.001\) \newline
Trial in block, \(\chi^2(2)=16.13\), \(p<.001\) & .178, .772 \\
Sickness symptom score & Gaussian LMM &
No interface effect reached significance &
Block order, \(\chi^2(3)=15.12\), \(p=.002\) & .022, .826 \\
System usability & Gaussian LMM &
Locomotion, \(\chi^2(1)=12.39\), \(p<.001\) \newline
Camera control, \(\chi^2(1)=78.72\), \(p<.001\) \newline
Locomotion by camera control, \(\chi^2(1)=12.57\), \(p<.001\) &
Block order, \(\chi^2(3)=8.14\), \(p=.043\) \newline
Trial in block, \(\chi^2(2)=8.85\), \(p=.012\) & .148, .763 \\
Technology acceptance & Gaussian LMM &
Locomotion, \(\chi^2(1)=3.96\), \(p=.047\) \newline
Camera control, \(\chi^2(1)=58.60\), \(p<.001\) \newline
Locomotion by camera control, \(\chi^2(1)=5.87\), \(p=.015\) &
Block order, \(\chi^2(3)=9.09\), \(p=.028\) & .124, .706 \\
Flow & Gaussian LMM &
Locomotion, \(\chi^2(1)=7.04\), \(p=.008\) \newline
Camera control, \(\chi^2(1)=27.51\), \(p<.001\) \newline
Locomotion by camera control, \(\chi^2(1)=12.83\), \(p<.001\) &
Block order, \(\chi^2(3)=16.71\), \(p<.001\) & .071, .788 \\
Camera switching & Negative binomial GLMM, log &
Camera control, \(\chi^2(1)=57.06\), \(p<.001\) &
Block order, \(\chi^2(3)=12.97\), \(p=.005\) \newline
Trial in block, \(\chi^2(2)=6.17\), \(p=.046\) & .310, .894 \\
PTC dwell allocation & Gaussian LMM &
Camera control, \(\chi^2(1)=38.49\), \(p<.001\) &
No task or order effect reached significance & .104, .533 \\
Section level PTC dwell & Gaussian LMM &
Condition, \(\chi^2(3)=37.18\), \(p<.001\) \newline
Condition by maze section, \(\chi^2(12)=40.71\), \(p<.001\) &
Maze section, \(\chi^2(4)=18.96\), \(p<.001\) \newline
Other task and order controls did not reach significance & .10, .86 \\
\bottomrule
\end{tabular}

\end{table*}

\subsection{Objective performance}

\subsubsection{Completion time}
Completion time was modeled with a Gamma GLMM using a log link. The omnibus model showed significant effects of locomotion, camera control, block order, and trial in block, while maze, target count, and the locomotion by camera-control interaction were not significant (Table~\ref{tab:model-summary}).

\begin{figure}[t]
  \centering
  \begin{minipage}[t]{0.49\linewidth}
    \centering
    \includegraphics[width=\linewidth]{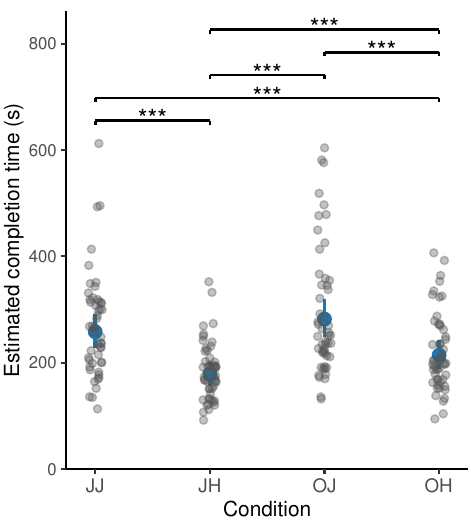}
  \end{minipage}\hfill
  \begin{minipage}[t]{0.49\linewidth}
    \centering
    \includegraphics[width=\linewidth]{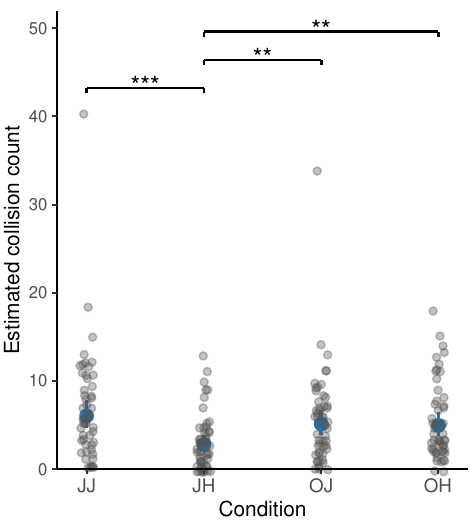}
  \end{minipage}
  \caption{Estimated marginal means (EMMs) by condition for (left) completion time and (right) wall-contact count. Error bars show 95\% CI and gray points show raw trials. Asterisks indicate Bonferroni-corrected significance (* $p<.05$, ** $p<.01$, *** $p<.001$).}
  \label{fig:completion-and-collision-condition}
  \label{fig:completion-time-condition}
  \label{fig:collision-count-condition}
\end{figure}

Estimated marginal means on the response scale indicated that completion time was shortest in \textit{JH} ($M=178.49$~s, $SE=11.40$), followed by \textit{OH} ($M=214.45$~s, $SE=13.69$), \textit{JJ} ($M=258.25$~s, $SE=16.49$), and \textit{OJ} ($M=282.36$~s, $SE=18.03$), see Fig.~\ref{fig:completion-time-condition}. Bonferroni-corrected post hoc comparisons showed that \textit{JH} was faster than \textit{JJ}, \textit{OJ}, and \textit{OH} (all $p<.001$), and that \textit{OH} was faster than \textit{JJ} ($p<.001$) and \textit{OJ} ($p<.001$). \textit{JJ} and \textit{OJ} did not differ ($p=.190$).

Order effects were evident. Estimated completion times were lower in Block~3 ($M=219.36$~s, $SE=14.00$) and Block~4 ($M=204.14$~s, $SE=13.04$) than in Block~1 ($M=249.01$~s, $SE=15.90$) and Block~2 ($M=250.33$~s, $SE=15.98$). Bonferroni-corrected comparisons showed that Block~1 was slower than Block~3 ($p=.014$) and Block~4 ($p<.001$), and Block~2 was slower than Block~3 ($p=.009$) and Block~4 ($p<.001$). Block~1 and Block~2, and Block~3 and Block~4 did not differ. This pattern indicates that participants became faster after the first two condition blocks. Within each condition block, Trial~1 ($M=250.11$~s, $SE=15.56$) was slower than Trial~2 ($M=224.19$~s, $SE=13.93$; $p=.007$) and Trial~3 ($M=216.58$~s, $SE=13.50$; $p<.001$), whereas Trial~2 and Trial~3 did not differ ($p=1.0$).

Taken together, completion time shows a clear advantage for head-guided camera control together with a strong learning effect across blocks and within the first trial of each block.

\subsubsection{Targets Captured}
Targets captured was analyzed using a generalized linear mixed-effects model (GLMM) with a grouped-binomial response and logit link. The outcome was modeled as captured targets out of required targets after capping overcapture trials at the required total for estimation. Candidate binomial and beta-binomial mixed models were compared using AIC, and the additive binomial model provided the best fit. Performance remained near ceiling in all four conditions, ranging from 98.80\% in JH to 99.78\% in OJ (Table~\ref{tab:objective-descriptives}).

No fixed effect reached significance in the target-completion model, and no Bonferroni-corrected pairwise condition contrast was significant (Table~\ref{tab:model-summary}). Participants captured nearly all required targets in every condition, leaving little evidence for reliable condition-related differences in target-capture success.

\begin{table}[t]
\centering
\footnotesize
\setlength{\tabcolsep}{2pt}
\caption{Model-based estimated marginal means (EMMs) on the response scale from the final mixed models for the objective and validated subjective outcomes.}
\label{tab:objective-descriptives}
\label{tab:subjective-descriptives}
\begin{tabular}{@{}p{0.36\columnwidth}C{0.13\columnwidth}C{0.13\columnwidth}C{0.13\columnwidth}C{0.13\columnwidth}@{}}
\toprule
Measure & JJ & JH & OJ & OH \\
\midrule
\multicolumn{5}{l}{\textbf{Objective outcomes}} \\
Completion time (s) & 258.25 & 178.49 & 282.36 & 214.45 \\
Targets captured (\%) & 99.32 & 98.80 & 99.78 & 99.61 \\
Wall-contact count & 6.05 & 2.79 & 5.13 & 4.96 \\
Wall-contact duration (s) & 19.49 & 11.59 & 23.91 & 14.22 \\
\midrule
\multicolumn{5}{l}{ \textbf{Questionnaire outcomes}} \\
NASA-TLX & 4.25 & 3.76 & 4.61 & 4.12 \\
Presence (PQ) & 5.50 & 7.10 & 6.63 & 7.27 \\
Sickness symptom score & 0.159 & 0.155 & 0.167 & 0.163 \\
System usability (SUS) & 58.93 & 76.96 & 69.17 & 76.95 \\
Technology acceptance (TAM) & 6.03 & 7.74 & 6.78 & 7.67 \\
Flow & 6.56 & 7.61 & 7.30 & 7.50 \\
\bottomrule
\end{tabular}
\end{table}

\subsubsection{Wall contact}
Wall-contact count was analyzed using a generalized linear mixed-effects model (GLMM) with a negative-binomial distribution (\texttt{nbinom1}) and log link. Candidate Poisson and negative-binomial mixed models were compared by AIC, and the \texttt{nbinom1} interaction model provided the best fit.

The wall-contact count model showed significant effects of locomotion, camera control, block order, and the locomotion by camera-control interaction. Maze, target count, and trial in block were not significant (Table~\ref{tab:model-summary}).

Estimated marginal means on the response scale indicated that collision count was lowest in \textit{JH} ($M=2.79$, $SE=0.46$), followed by \textit{OH} ($M=4.96$, $SE=0.70$), \textit{OJ} ($M=5.13$, $SE=0.71$), and \textit{JJ} ($M=6.05$, $SE=0.80$), see Fig.~\ref{fig:collision-count-condition}. Bonferroni-corrected post hoc comparisons showed that \textit{JH} yielded fewer collisions than \textit{JJ} ($p<.001$), \textit{OJ} ($p=.001$), and \textit{OH} ($p=.002$). The remaining condition contrasts were not significant.

Estimated counts decreased across condition blocks from Block~1 ($M=5.73$, $SE=0.78$) to Block~4 ($M=4.05$, $SE=0.59$), with Block~2 ($M=4.42$, $SE=0.64$) and Block~3 ($M=4.20$, $SE=0.61$) in between. However, no Bonferroni-corrected pairwise block-order contrasts remained significant, indicating a gradual downward trend rather than one isolated contrast. 
Overall, the count model indicates that head-guided camera control, especially when paired with joystick locomotion, was associated with fewer collisions than the camera-joystick conditions.

Wall-contact duration was modeled with a Tweedie GLMM. The model showed significant effects of camera control and block order, while locomotion, maze, target count, and trial in block were not significant (Table~\ref{tab:model-summary}).

Collision duration was lowest in \textit{JH} ($M=11.59$~s, $SE=2.10$), followed by \textit{OH} ($M=14.22$~s, $SE=2.51$), \textit{JJ} ($M=19.49$~s, $SE=3.38$), and \textit{OJ} ($M=23.91$~s, $SE=4.12$), as summarized in Table~\ref{tab:objective-descriptives}. Bonferroni-corrected comparisons showed that \textit{JH} had lower collision duration than \textit{JJ} ($p<.001$) and \textit{OJ} ($p<.001$), and that \textit{OH} also had lower collision duration than \textit{OJ} ($p<.001$). This mirrors the count result: head-guided camera control reduced both how often participants contacted walls and how long those contacts lasted. Collision duration also decreased across condition blocks: Block~1 was higher than Block~2 ($p=.013$), Block~3 ($p=.043$), and Block~4 ($p=.012$), whereas the differences among Blocks~2--4 were not significant. A sensitivity refit changed the significance of locomotion and trial in block; interpretation therefore focuses on the stable effects of camera control, block order, and the condition-level contrasts.

\subsection{Subjective measures}

All questionnaire outcomes were analyzed with Gaussian linear mixed-effects models, and condition means are summarized in Table~\ref{tab:subjective-descriptives}. NASA-TLX showed significant effects of locomotion, camera control, block order, and trial in block. Workload was lowest in \textit{JH} ($M=3.76$, $SE=0.27$) and highest in \textit{OJ} ($M=4.61$, $SE=0.27$). The sickness symptom score showed no reliable condition effect and remained low across all interfaces (range $=0.155$--$0.167$).

Presence, flow, system usability, and technology acceptance all showed significant camera-control effects. Presence, flow, usability, and technology acceptance also showed locomotion by camera-control interactions. Across these scales, \textit{JJ} was consistently weakest, both head-guided conditions were highest or near-highest, and \textit{OJ} was generally intermediate. Presence was highest in \textit{OH} ($M=7.27$, $SE=0.41$) and \textit{JH} ($M=7.10$, $SE=0.41$). Flow was highest in \textit{JH} ($M=7.61$, $SE=0.41$) and \textit{OH} ($M=7.50$, $SE=0.41$), while usability and technology acceptance were highest in the two head-guided conditions (SUS: both $M\approx76.95$; TAM: JH $M=7.74$, OH $M=7.67$). Together, the questionnaire measures closely matched the objective pattern: head-guided camera control improved the felt quality of remote operation, while \textit{OH} provided the strongest experiential compromise.

\subsection{Camera behavior}

\subsubsection{Camera switching frequency}
The following camera-behavior outcomes were analyzed exploratorily to help interpret the condition differences in performance and experience. Camera switching frequency was modeled with a negative-binomial GLMM. The model showed significant effects of camera control, block order, and trial in block. Locomotion, maze, and target count were not significant (Table~\ref{tab:model-summary}).

\begin{figure}[t]
  \centering
  \begin{minipage}[t]{0.49\linewidth}
    \centering
    \includegraphics[width=\linewidth]{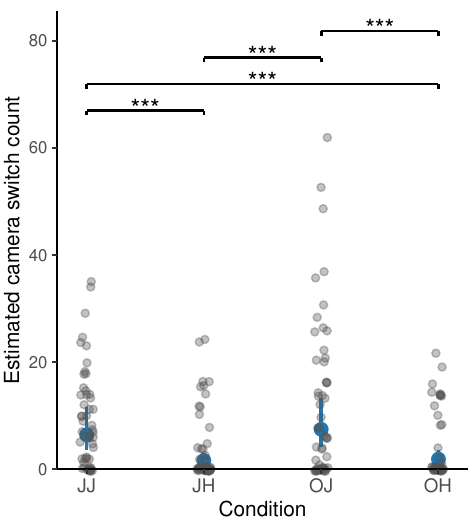}
  \end{minipage}\hfill
  \begin{minipage}[t]{0.49\linewidth}
    \centering
    \includegraphics[width=\linewidth]{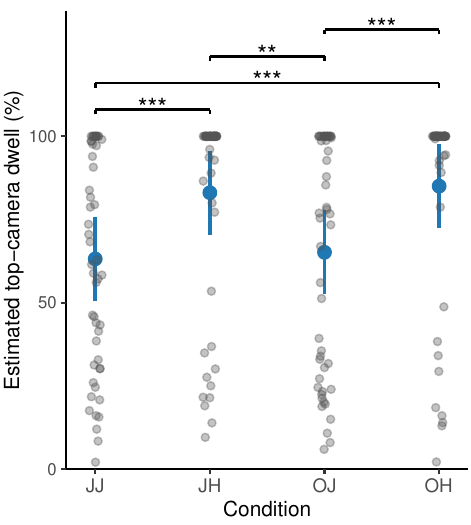}
  \end{minipage}
  \caption{Estimated marginal means (EMMs) by condition for (left) camera switching frequency and (right) camera dwell allocation, expressed as the percentage of trial time spent in the PTC view. The RBC percentage is the complement of the PTC quantity. Error bars show 95\% confidence intervals and gray points show raw trials. Asterisks indicate Bonferroni-corrected pairwise differences (* $p<.05$, ** $p<.01$, *** $p<.001$).}
  \label{fig:camera-switch-and-dwell-condition}
  \label{fig:camera-switch-condition}
  \label{fig:camera-dwell-condition}
\end{figure}

Estimated marginal means indicated that switching was least frequent in the head-guided camera conditions, with \textit{JH} showing the fewest switches ($M=1.62$, $SE=0.53$) and \textit{OH} remaining similarly low ($M=1.88$, $SE=0.61$). Both joystick-camera conditions showed substantially higher switching, with \textit{JJ} at $M=6.42$ ($SE=1.93$) and \textit{OJ} at $M=7.45$ ($SE=2.18$), see Fig.~\ref{fig:camera-switch-condition}. Bonferroni-adjusted post hoc comparisons showed that both head-guided camera conditions differed from both joystick-camera conditions, whereas \textit{JJ} and \textit{OJ} did not differ, and \textit{JH} and \textit{OH} did not differ. This pattern shows that head-guided camera control reduced camera switching across both locomotion modes.

Switching frequency also declined across condition blocks, from Block~1 ($M=4.63$, $SE=1.45$) to Block~3 ($M=2.16$, $SE=0.72$), with Block~2 ($M=4.26$, $SE=1.35$) and Block~4 ($M=3.42$, $SE=1.09$) in between. Bonferroni-adjusted tests showed two significant block-order contrasts. Within each condition block, switching was highest in Trial~1 ($M=4.35$, $SE=1.31$) and decreased to Trial~2 ($M=3.56$, $SE=1.08$) and Trial~3 ($M=2.71$, $SE=0.85$), with one Bonferroni-adjusted pairwise contrast significant. Sensitivity refits excluding flagged trials changed only the significance of target count and did not alter the main conclusion that head-guided camera control substantially reduced switching frequency.

\subsubsection{Camera dwell allocation}
Camera dwell allocation was analyzed as the percentage of trial time spent in the PTC view, with the RBC percentage interpreted as its complement. Inferential testing used the PTC share because the two camera shares sum to 100\%. The model showed a significant effect of camera control, while locomotion, maze, target count, block order, and trial in block were not significant (Table~\ref{tab:model-summary}).

Estimated marginal means showed greater PTC dwell in the head-guided camera conditions. \textit{JH} spent $M=83.05\%$ ($SE=6.13$) of trial time in the PTC view and \textit{OH} spent $M=85.02\%$ ($SE=6.13$), leaving corresponding RBC shares of roughly $17$--$15\%$. The joystick-camera conditions showed more mixed allocation, with PTC dwell at \textit{JJ} $M=63.16\%$ ($SE=6.13$) and \textit{OJ} $M=65.14\%$ ($SE=6.13$), corresponding to RBC shares of roughly $37$--$35\%$ (Fig.~\ref{fig:camera-dwell-condition}).

Bonferroni-adjusted pairwise contrasts showed that both head-guided camera conditions differed from both joystick-camera conditions. \textit{JJ} and \textit{OJ} did not differ, and \textit{JH} and \textit{OH} did not differ. This pattern indicates that head-guided camera control shifted viewing toward sustained use of the PTC and reduced reliance on the RBC across both locomotion modes. Absolute viewing-time descriptives followed the same direction, but relative allocation is more interpretable because viewing time changes with overall trial duration. Sensitivity refits excluding flagged trials did not change the omnibus conclusions.

The target-capture screenshots showed a similar viewing preference. The share of captures taken from the PTC was higher in the head-guided conditions (\textit{JH}: $M=93.34\%$, \textit{OH}: $M=94.17\%$) than in the joystick-camera conditions (\textit{JJ}: $M=81.74\%$, \textit{OJ}: $M=83.77\%$). This model showed a significant main effect of camera control and no locomotion effect.

Use of both cameras within the same trial showed the opposite pattern. Trials were much more likely to use both cameras in the joystick-camera conditions (\textit{JJ}: $M=94.01\%$, \textit{OJ}: $M=90.28\%$) than in the head-guided-camera conditions (\textit{JH}: $M=25.23\%$, \textit{OH}: $M=16.64\%$). This model also showed a significant main effect of camera control and no locomotion effect. Together, these outcomes indicate that head-guided camera control reduced continued view negotiation during both navigation and target capture.

\subsubsection{Camera-use strategy across maze sections}
Camera dwell allocation was analyzed at the section level using a Gaussian linear mixed-effects model (LMM). Each trial was divided into five ordered maze sections from start to end using the maze-boundary checkpoints. For interpretation, Sections~1--2 are treated as the beginning of the maze, Section~3 as the middle, and Sections~4--5 as the end. Relative to the general model structure described above, these section-level models added \textit{maze section}; for dwell allocation, the \textit{condition $\times$ maze-section} interaction was also tested. Random intercepts for \textit{participant} and \textit{trial} accounted for repeated section rows within trials.

The section-level dwell model showed significant effects of condition, maze section, and their interaction. Maze, target count, block order, and trial in block were not significant (Table~\ref{tab:model-summary}).

Collapsing across sections, PTC share was lower in the joystick-camera conditions than in the head-guided-camera conditions. The estimated means were JJ $M=62.39\%$ ($SE=6.54$), OJ $M=64.34\%$ ($SE=6.56$), JH $M=81.99\%$ ($SE=6.56$), and OH $M=85.30\%$ ($SE=6.51$). Bonferroni-adjusted post hoc tests showed that both head-guided camera conditions differed from both joystick-camera conditions, whereas the two joystick-camera conditions did not differ from each other and the two head-guided camera conditions did not differ from each other.

Across the route, the head-guided camera conditions kept PTC dwell high and stable. JH changed only slightly from Section~1 ($80.8\%$) to Section~5 ($79.8\%$), and OH remained similarly high from Section~1 ($84.5\%$) to Section~5 ($87.0\%$). The joystick-camera conditions showed lower PTC allocation and a stronger late-maze shift toward the RBC, especially in JJ, where PTC dwell decreased from $70.6\%$ in Section~1 to $55.3\%$ in Section~5. Sensitivity refits excluding flagged section rows did not change the omnibus significance pattern.

A complementary section-level switching model supported the same interpretation. Collapsing across sections, switching was lower in the head-guided camera conditions (JH $M=0.13$, OH $M=0.17$) than in the joystick-camera conditions (JJ $M=0.80$, OJ $M=0.84$). Descriptively, switching remained especially elevated in the middle and later maze sections for OJ, whereas both head-guided camera conditions stayed low throughout. Together, the section-level dwell and switching results indicate that HGC supported a stable camera strategy across the route, while joystick-based camera control continued to demand view management deeper into the maze.

\subsubsection{Exploratory camera-strategy models}

We also examined whether camera-use strategy was consistent with part of the condition effects on performance. Each primary performance model was compared with an augmented model that added total section switches, whether the last switch occurred in the late maze, and mean late-maze PTC dwell. These exploratory models reduced the omnibus condition effect by $31.7\%$ for completion time, $28.9\%$ for wall-contact count, and $57.8\%$ for wall-contact duration.

Across the three models, more switching predicted slower completion, more collisions, and longer collision duration. Late-maze PTC dwell predicted fewer collisions, whereas late switching predicted slower completion. These analyses do not establish causal mediation, but they support the interpretation that stable camera-use strategy was associated with part of the advantage of the head-guided conditions.

\section{Discussion}
Teleoperation work treats viewpoint design as part of the control loop because camera and motion control mappings change what the operator can attend to, how quickly the remote scene can be understood, and how much effort is spent maintaining a usable view \cite{doisy2017comparison,Lin2023PerceptionMotion,Liu2025CameraControl,Yoon2025ViewpointControl}. The present study extends this argument to a task where navigation, inspection, and visual documentation had to happen together. The main result is not that one input device was uniformly better than another. Instead, the task exposed where coordination costs entered immersive robot telepresence. Viewpoint control shaped performance and experience more consistently than locomotion modality, which shows that the camera mapping was part of the task structure rather than a secondary display choice.

This contribution becomes clearer in the camera behavior results. Accounts of transparent telepresence emphasize that control input and visual feedback must remain easy to relate during action, otherwise the interface itself becomes part of the task \cite{Toet2020Embodiment,Aguilar2025EmbodiedTelepresence}. In this study, HGC changed how participants used the available views and reduced the need for repeated view negotiation. In a navigation and inspection task, every camera transition adds a small interpretive cost. The operator must judge whether the current view is sufficient, switch if needed, recover the scene from the new viewpoint, and then continue moving or inspecting. The section-level results and exploratory strategy models point in the same direction. Viewpoint transition work is useful here because switching views is a control decision that can interrupt how the operator understands the current scene and prepares the next action \cite{Yoon2025Transitions}. HGC appears to have reduced this interruption by supporting a more stable viewing strategy across the route.

The locomotion findings are best read through the same task fit logic. VR locomotion research has shown that more body involvement can support experience and spatial updating, but it can also introduce balance, regulation, learning, or control demands when the wider task does not fit the method \cite{wang2024investigating,wehden2021slippery}. Our results follow that tradeoff. Omnidirectional treadmill locomotion did not reliably improve task efficiency, yet OH produced favorable experiential ratings and a stable camera-use pattern. This suggests that body coupled locomotion can improve the felt quality of remote operation when it is paired with low-overhead viewpoint control, but it cannot compensate for a demanding camera mapping.

The quadruped platform clarifies the scope of this telepresence claim. Telepresence accounts that emphasize agency, self-location, and transparent action through a remote system do not require the design argument to rest only on humanoid morphology \cite{Toet2020Embodiment,Nostadt2020Embodiment,Aguilar2025EmbodiedTelepresence}. Here, the robot body, gait, camera height, and motion limits differed from the operator's body, and both head motion and treadmill motion were mapped to bounded robot mechanisms. The claim is therefore limited to partial body coupled telepresence. Even so, the interface provided task relevant body cues. The RBC supported heading and wall-clearance judgments, the PTC could return to a forward reference, and looking downward preserved visibility of the robot body width. What mattered was whether the operator could connect their control actions to useful remote visual feedback during movement and inspection.

Several limits define the scope of these conclusions. The sample size was fixed at 17 participants, which limits sensitivity to smaller effects. Target completion was near ceiling, so that outcome could not strongly distinguish the four interfaces. The task was a controlled laboratory navigation and inspection scenario with foam-wall contacts under low speed supervision. Practical systems would need collision avoidance, speed adaptation, shared autonomy, or locomotion assistance for less controlled spaces. Glass to glass visual latency was not measured, although the system logged round-trip timing information. The four conditions also combined posture, body coupling, and controller allocation, and seated joystick or swivel chair operation may be more realistic for some deployments. Treadmill based telepresence adds physical effort and accessibility constraints. Finally, the camera strategy models are exploratory and support a mechanism interpretation, but they do not establish causal mediation.

The practical implication follows from both previous camera-control work and the present robot task. Prior studies connect camera control and viewpoint transitions to workload, attention, and remote action quality \cite{Lin2023PerceptionMotion,Liu2025CameraControl,Yoon2025ViewpointControl}. Our results show that this connection remains critical when the operator is also using a body coupled locomotion interface and controlling a real quadruped robot. For navigation and inspection tasks, omnidirectional locomotion therefore needs a camera-control method that keeps visual inspection easy to maintain while moving.

\section{Conclusion}
In this immersive robot telepresence task, the way the remote viewpoint was coupled to the user mattered more consistently than locomotion modality alone. Head-guided camera control reduced camera-management overhead, supported a stable viewing strategy, and was associated with better objective performance and more favorable subjective experience. These effects emerged on a quadruped robot, suggesting that effective immersive telepresence depends less on morphology alone than on how naturally user actions couple to useful remote visual feedback. For telepresence design, viewpoint-control mapping is a central design decision, and body coupled locomotion is most promising when paired with equally low-overhead viewpoint control.

\bibliographystyle{IEEEtran}

\bibliography{Bibliography}
\end{document}


\maketitle

\begin{abstract}
This supplement provides the supporting model results, practice effects, camera-behavior analyses, camera-strategy augmented models, and exploratory questionnaire summaries for the main manuscript. The sections are organized around objective performance, camera use, validated questionnaires, and exploratory post-trial responses.
\end{abstract}

\begin{IEEEkeywords}
Robot telepresence, immersive teleoperation, viewpoint control, supplementary material.
\end{IEEEkeywords}

\section{Supplementary Analyses}
The main manuscript reports the primary objective, subjective, and camera-behavior findings. This supplement retains the detailed statistical reporting for supporting analyses and groups the results by their role in the study. Objective performance analyses first document practice and order effects. Camera-behavior analyses then show how participants managed the two available views. The camera-strategy augmented models examine whether these viewing behaviors account for part of the condition effects. The final sections report validated questionnaire outcomes and exploratory post-trial and post-study responses.

\subsection{Objective Performance and Practice Effects}
These analyses complement the main performance results by showing how completion time and wall contact changed across condition order and repeated trials. The same model families and reporting conventions used in the main manuscript are retained here.

\begin{table*}[t]
\centering
\scriptsize
\setlength{\tabcolsep}{2.5pt}
\renewcommand{\arraystretch}{1.08}
\caption{Supplementary objective-performance model details and practice effects. Effect entries report \(\chi^2(df)\), \(p\). \(R^2_m\) and \(R^2_c\) denote marginal and conditional \(R^2\).}
\label{tab:supp-objective-practice}
\begin{tabular}{@{}L{0.15\textwidth}L{0.17\textwidth}L{0.22\textwidth}L{0.30\textwidth}L{0.10\textwidth}@{}}
\toprule
Outcome & Model & Interface effects & Task, order, and practice effects & \(R^2_m\) / \(R^2_c\) \\
\midrule
Completion time & Gamma GLMM, log &
Locomotion, \(\chi^2(1)=21.51\), \(p<.001\) \newline
Camera control, \(\chi^2(1)=119.96\), \(p<.001\) \newline
Locomotion by camera control, \(\chi^2(1)=2.58\), \(p=.108\) &
Maze, \(\chi^2(2)=1.87\), \(p=.392\) \newline
Target count, \(\chi^2(2)=1.55\), \(p=.460\) \newline
Block order, \(\chi^2(3)=34.26\), \(p<.001\) \newline
Trial in block, \(\chi^2(2)=16.91\), \(p<.001\) \newline
Block~1 and Block~2 were slower than Block~3 and Block~4. Trial~1 was slower than Trial~2 and Trial~3. & .304 / .696 \\
\midrule
Wall-contact count & Negative-binomial GLMM, log &
Locomotion, \(\chi^2(1)=3.87\), \(p=.049\) \newline
Camera control, \(\chi^2(1)=14.95\), \(p<.001\) \newline
Locomotion by camera control, \(\chi^2(1)=12.64\), \(p<.001\) &
Maze, \(\chi^2(2)=3.13\), \(p=.209\) \newline
Target count, \(\chi^2(2)=3.62\), \(p=.164\) \newline
Block order, \(\chi^2(3)=8.23\), \(p=.041\) \newline
Trial in block, \(\chi^2(2)=3.52\), \(p=.172\) \newline
Estimated counts decreased from Block~1 to Block~4, but no Bonferroni-corrected block contrast remained significant. & .161 / .348 \\
\midrule
Wall-contact duration & Tweedie GLMM, log &
Locomotion, \(\chi^2(1)=2.61\), \(p=.106\) \newline
Camera control, \(\chi^2(1)=16.92\), \(p<.001\) &
Maze, \(\chi^2(2)=3.30\), \(p=.192\) \newline
Target count, \(\chi^2(2)=1.06\), \(p=.590\) \newline
Block order, \(\chi^2(3)=14.00\), \(p=.003\) \newline
Trial in block, \(\chi^2(2)=5.55\), \(p=.062\) \newline
Block~1 had longer contact duration than Blocks~2, 3, and 4. & .171 / .473 \\
\bottomrule
\end{tabular}
\end{table*}

\subsubsection{Completion time}
Completion time was analyzed using a generalized linear mixed-effects model (GLMM) with a Gamma distribution and log link. Candidate Gaussian and Gamma mixed models were compared, and the Gamma(log) model provided the best fit. Table~\ref{tab:supp-objective-practice} summarizes the fixed-effect tests and model \(R^2\) values.

\begin{figure}[!t]
  \centering
  \includegraphics[width=\linewidth]{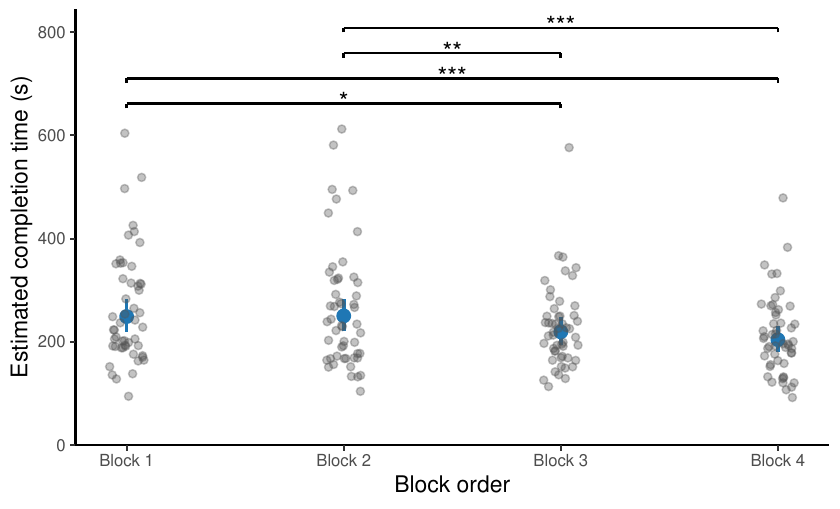}
  \caption{Estimated marginal means (EMMs) of completion time by block order from the Gamma GLMM (log link). Error bars show 95\% confidence intervals and gray points show raw trials. Asterisks indicate Bonferroni-corrected pairwise differences (* $p<.05$, ** $p<.01$, *** $p<.001$).}
  \label{fig:completion-time-block-order}
\end{figure}

\begin{figure}[!t]
  \centering
  \includegraphics[width=\linewidth]{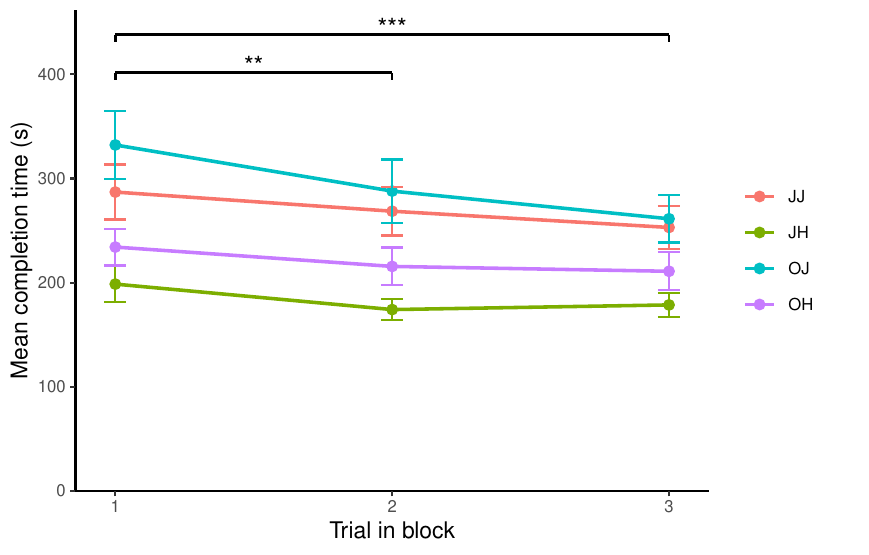}
  \caption{Estimated marginal means (EMMs) of completion time by trial in block from the Gamma GLMM (log link), shown separately for the four interface conditions. Error bars show 95\% confidence intervals and gray points show raw trials. Asterisks indicate Bonferroni-corrected pairwise differences (* $p<.05$, ** $p<.01$, *** $p<.001$).}
  \label{fig:completion-time-learning}
\end{figure}

Estimated completion times were lower in Block~3 ($M=219.36$~s, $SE=14.00$) and Block~4 ($M=204.14$~s, $SE=13.04$) than in Block~1 ($M=249.01$~s, $SE=15.90$) and Block~2 ($M=250.33$~s, $SE=15.98$), see Fig.~\ref{fig:completion-time-block-order}. Bonferroni-corrected comparisons showed that Block~1 was slower than Block~3 ($p=.014$) and Block~4 ($p<.001$), and Block~2 was slower than Block~3 ($p=.009$) and Block~4 ($p<.001$). Block~1 and Block~2 did not differ, and Block~3 and Block~4 did not differ. This pattern indicates that participants became faster after the first two condition blocks and then stabilized.

Within each condition block, Trial~1 ($M=250.11$~s, $SE=15.56$) was slower than Trial~2 ($M=224.19$~s, $SE=13.93$; $p=.007$) and Trial~3 ($M=216.58$~s, $SE=13.50$; $p<.001$), whereas Trial~2 and Trial~3 did not differ ($p=1.000$), see Fig.~\ref{fig:completion-time-learning}. Taken together, completion time shows a clear advantage for head-guided camera control together with a strong learning effect across blocks and within the first trial of each block.

\subsubsection{Wall-contact Count and Duration}
\textit{Wall-contact count.} Wall-contact count was analyzed using a generalized linear mixed-effects model (GLMM) with a negative-binomial distribution (\texttt{nbinom1}) and log link. Candidate Poisson and negative-binomial mixed models were compared, and the \texttt{nbinom1} interaction model provided the best fit. Table~\ref{tab:supp-objective-practice} summarizes the fixed-effect tests and model \(R^2\) values.

\begin{figure}[!t]
  \centering
  \includegraphics[width=\linewidth]{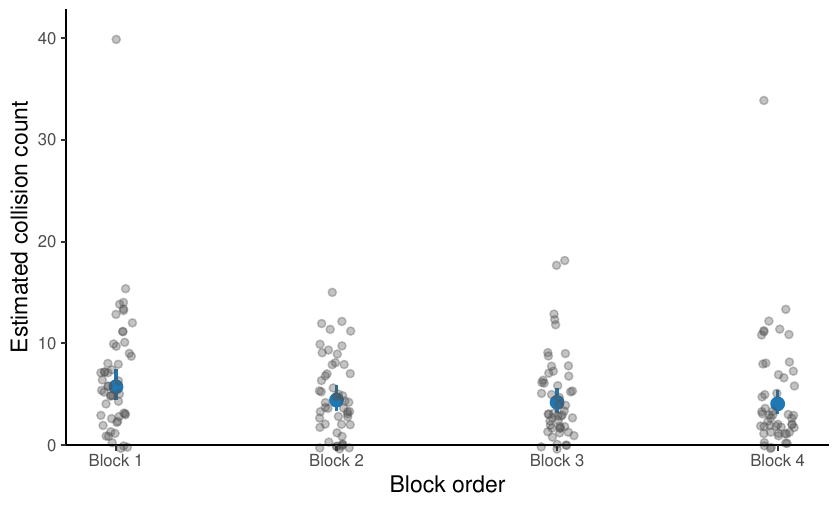}
  \caption{Estimated marginal means (EMMs) of wall-contact count by block order from the negative-binomial GLMM (log link). Error bars show 95\% confidence intervals and gray points show raw trials. Asterisks indicate Bonferroni-corrected pairwise differences when present.}
  \label{fig:collision-count-block-order}
\end{figure}

\begin{figure}[!t]
  \centering
  \includegraphics[width=\linewidth]{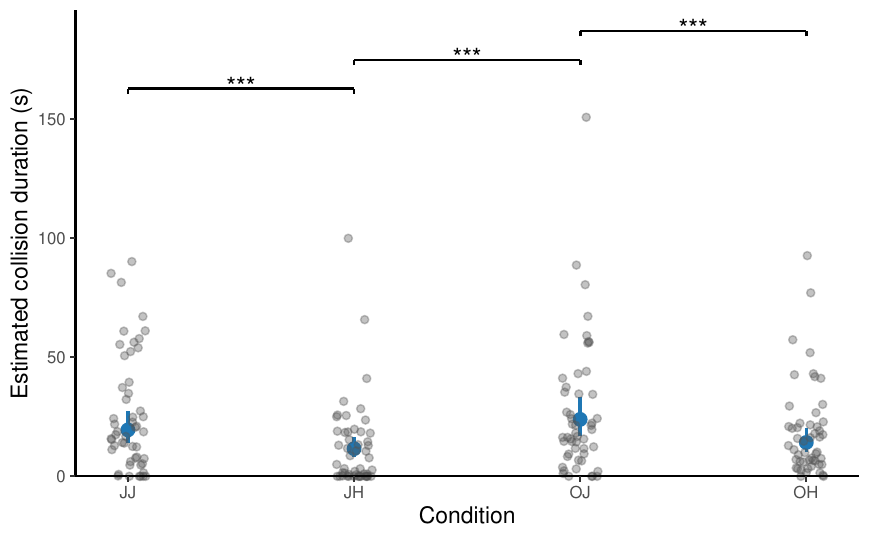}
  \caption{Estimated marginal means (EMMs) of wall-contact duration by condition from the Tweedie GLMM (log link). Error bars show 95\% confidence intervals and gray points show raw trials. Asterisks indicate Bonferroni-corrected pairwise differences (* $p<.05$, ** $p<.01$, *** $p<.001$).}
  \label{fig:collision-duration-condition}
\end{figure}

\begin{figure}[!t]
  \centering
  \includegraphics[width=\linewidth]{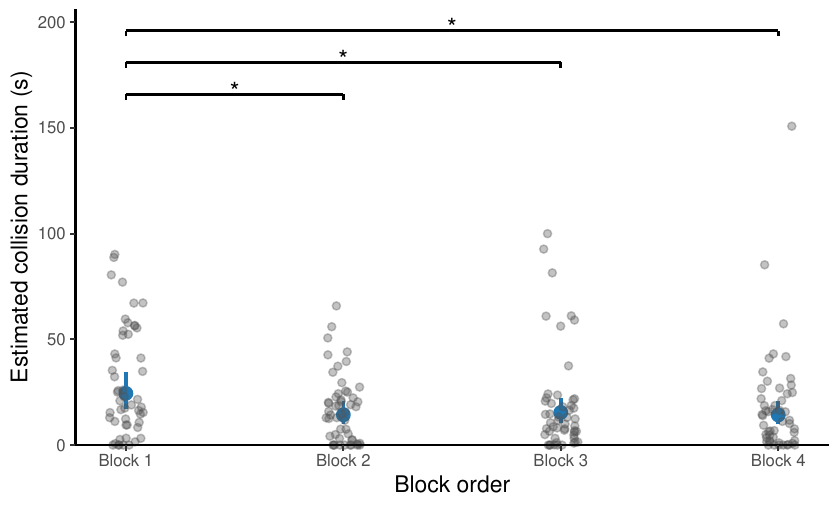}
  \caption{Estimated marginal means (EMMs) of wall-contact duration by block order from the Tweedie GLMM (log link). Error bars show 95\% confidence intervals and gray points show raw trials. Asterisks indicate Bonferroni-corrected pairwise differences (* $p<.05$, ** $p<.01$, *** $p<.001$).}
  \label{fig:collision-duration-block-order}
\end{figure}

Estimated wall-contact counts decreased from Block~1 ($M=5.73$, $SE=0.78$) to Block~4 ($M=4.05$, $SE=0.59$), with Block~2 ($M=4.42$, $SE=0.64$) and Block~3 ($M=4.20$, $SE=0.61$) in between, see Fig.~\ref{fig:collision-count-block-order}. However, no Bonferroni-corrected pairwise block-order contrasts remained significant, indicating a gradual downward trend rather than one isolated block contrast. A sensitivity refit excluding flagged trials did not change the omnibus significance pattern.

\textit{Wall-contact duration.} Wall-contact duration was analyzed with a Tweedie GLMM. Table~\ref{tab:supp-objective-practice} summarizes the fixed-effect tests and model \(R^2\) values.

Collision duration was lowest in JH ($M=11.59$~s, $SE=2.10$), followed by OH ($M=14.22$~s, $SE=2.51$), JJ ($M=19.49$~s, $SE=3.38$), and OJ ($M=23.91$~s, $SE=4.12$), see Fig.~\ref{fig:collision-duration-condition}. Bonferroni-corrected comparisons showed that JH had lower collision duration than JJ ($p<.001$) and OJ ($p<.001$), and that OH also had lower collision duration than OJ ($p<.001$). This mirrors the count result because head-guided camera control reduced both how often participants contacted walls and how long those contacts lasted.

Collision duration also decreased across condition blocks. Block~1 was higher than Block~2 ($p=.013$), Block~3 ($p=.043$), and Block~4 ($p=.012$), whereas the differences among Blocks~2--4 were not significant, see Fig.~\ref{fig:collision-duration-block-order}. A sensitivity refit changed the significance of locomotion and trial in block, so the more stable results were the effects of camera control, block order, and the condition-level contrasts.

\subsection{Camera Behavior and Viewpoint Management}
The camera-behavior analyses describe how participants used the Robot Body Camera and the Pan-Tilt Camera during navigation and target capture. These outcomes are useful because they connect the interface manipulation to the viewing strategies that emerged during the task.

\begin{table*}[t]
\centering
\scriptsize
\setlength{\tabcolsep}{2.3pt}
\renewcommand{\arraystretch}{1.08}
\caption{Supplementary camera-behavior model details. Effect entries report \(\chi^2(df)\), \(p\), when available in the analysis output.}
\label{tab:supp-camera-behavior}
\begin{tabular}{@{}L{0.14\textwidth}L{0.18\textwidth}L{0.24\textwidth}L{0.30\textwidth}L{0.08\textwidth}@{}}
\toprule
Outcome & Model & Interface effects & Task, order, and strategy details & \(R^2_m\) / \(R^2_c\) \\
\midrule
Camera switching frequency & Negative-binomial GLMM, log &
Locomotion, \(\chi^2(1)=0.85\), \(p=.358\) \newline
Camera control, \(\chi^2(1)=57.06\), \(p<.001\) &
Maze, \(\chi^2(2)=0.15\), \(p=.929\) \newline
Target count, \(\chi^2(2)=5.70\), \(p=.058\) \newline
Block order, \(\chi^2(3)=12.97\), \(p=.005\) \newline
Trial in block, \(\chi^2(2)=6.17\), \(p=.046\) & .310 / .894 \\
\midrule
Capture camera share & Beta-binomial GLMM, logit &
Locomotion, \(\chi^2(1)=0.35\), \(p=.556\) \newline
Camera control, \(\chi^2(1)=20.20\), \(p<.001\) &
Maze, \(\chi^2(2)=1.60\), \(p=.450\) \newline
Target count, \(\chi^2(2)=4.69\), \(p=.096\) \newline
Block order, \(\chi^2(3)=1.63\), \(p=.652\) \newline
Trial in block, \(\chi^2(2)=0.44\), \(p=.801\) & .208 / 1.000 \\
\midrule
Use of both cameras & Binomial GLMM, logit &
Locomotion, \(\chi^2(1)=1.16\), \(p=.281\) \newline
Camera control, \(\chi^2(1)=32.77\), \(p<.001\) &
Maze, \(\chi^2(2)=0.47\), \(p=.792\) \newline
Target count, \(\chi^2(2)=2.38\), \(p=.304\) \newline
Block order, \(\chi^2(3)=13.79\), \(p=.003\) \newline
Trial in block, \(\chi^2(2)=9.82\), \(p=.007\) & .314 / .814 \\
\midrule
Pan-tilt-camera viewing time & Gaussian LMM &
Locomotion, \(\chi^2(1)=10.76\), \(p=.001\) \newline
Camera control, \(\chi^2(1)=5.19\), \(p=.023\) &
Block order, \(\chi^2(3)=12.26\), \(p=.007\) \newline
Sensitivity refits changed the significance of camera control. & .069 / .610 \\
\midrule
Bottom-camera viewing time & Tweedie GLMM, log &
Camera control, \(\chi^2(1)=38.06\), \(p<.001\) \newline
Locomotion did not reach significance. &
Maze, target count, block order, and trial in block did not reach significance. \newline
Sensitivity refits did not change the omnibus conclusions. & .164 / .702 \\
\midrule
Section-level camera switching & Negative-binomial GLMM, log &
Condition, \(\chi^2(3)=65.47\), \(p<.001\) &
Maze section, \(\chi^2(4)=34.37\), \(p<.001\) \newline
Block order, \(\chi^2(3)=14.98\), \(p=.002\) \newline
Trial in block, \(\chi^2(2)=8.49\), \(p=.014\) \newline
Maze and target count did not reach significance. & .20 / .82 \\
\bottomrule
\end{tabular}
\end{table*}

\subsubsection{Camera switching frequency}
Camera switching frequency was analyzed using a negative-binomial GLMM with a log link. Table~\ref{tab:supp-camera-behavior} summarizes the fixed-effect tests and model \(R^2\) values. The figures below show the corresponding block-order and within-block patterns.

\begin{figure}[!t]
  \centering
  \includegraphics[width=\linewidth]{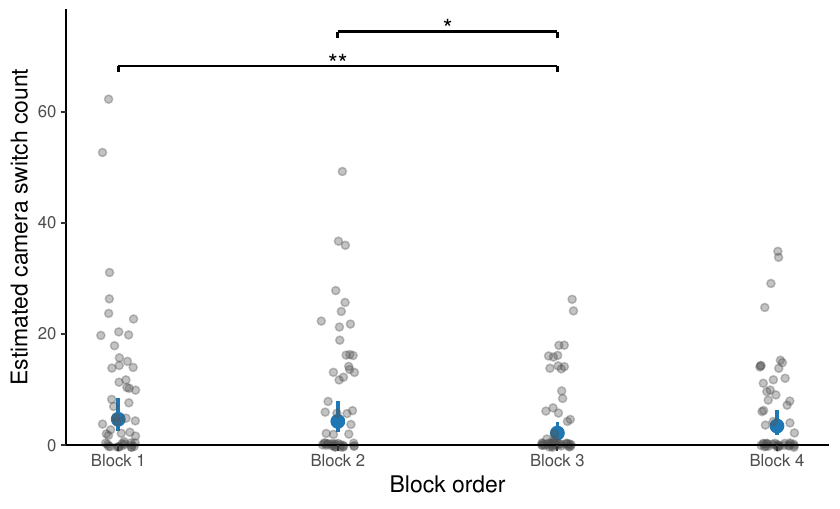}
  \caption{Estimated marginal means (EMMs) of camera switching frequency by block order from the negative-binomial GLMM (log link). Error bars show 95\% confidence intervals and gray points show raw trials. Asterisks indicate Bonferroni-corrected pairwise differences when present.}
  \label{fig:camera-switch-block-order}
\end{figure}

\begin{figure}[!t]
  \centering
  \includegraphics[width=\linewidth]{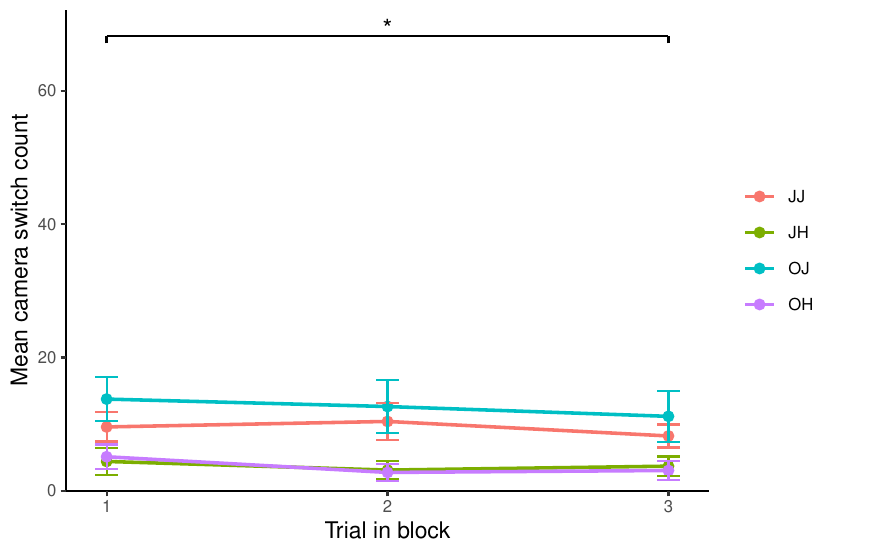}
  \caption{Estimated marginal means (EMMs) of camera switching frequency by trial in block from the negative-binomial GLMM (log link), shown separately for the four interface conditions. Error bars show 95\% confidence intervals and gray points show raw trials. Asterisks indicate Bonferroni-corrected pairwise differences when present.}
  \label{fig:camera-switch-learning}
\end{figure}

Estimated switching frequency was highest in Block~1 ($M=4.63$, $SE=1.45$) and lowest in Block~3 ($M=2.16$, $SE=0.72$), with Block~2 ($M=4.26$, $SE=1.35$) and Block~4 ($M=3.42$, $SE=1.09$) in between, see Fig.~\ref{fig:camera-switch-block-order}. Bonferroni-adjusted tests showed two significant block-order contrasts overall. Within each condition block, switching was highest in Trial~1 ($M=4.35$, $SE=1.31$), decreasing to Trial~2 ($M=3.56$, $SE=1.08$) and Trial~3 ($M=2.71$, $SE=0.85$), with one Bonferroni-adjusted pairwise contrast significant, see Fig.~\ref{fig:camera-switch-learning}. These plots show that switching decreased with practice, while the condition pattern reported in the main paper remained centered on lower switching under head-guided camera control. Sensitivity refits excluding flagged trials changed only the significance of target count and did not alter that conclusion.

\subsubsection{Capture camera share}
Capture camera share was analyzed using a generalized linear mixed-effects model (GLMM) with a beta-binomial distribution and logit link. The outcome was defined as the share of target-capture screenshots taken from the pan-tilt camera, with the bottom-camera share interpreted as its complement. Candidate binomial and beta-binomial mixed models were compared, and the additive beta-binomial model provided the best fit. Table~\ref{tab:supp-camera-behavior} summarizes the fixed-effect tests.

\begin{figure}[!t]
  \centering
  \includegraphics[width=\linewidth]{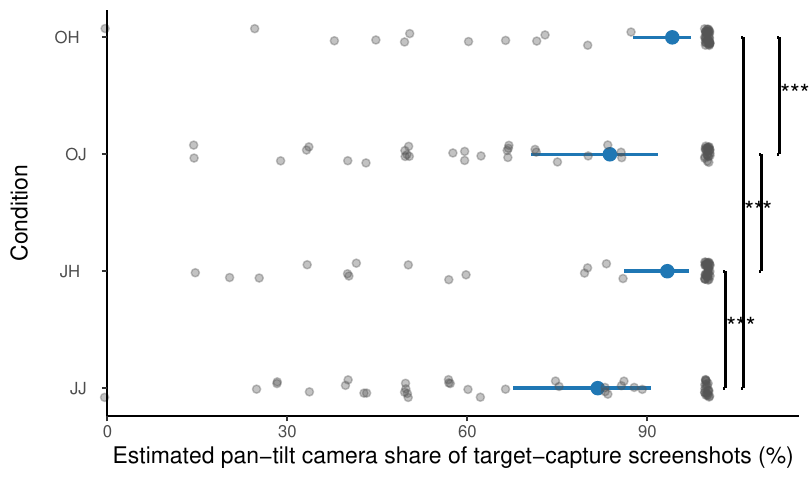}
  \caption{Estimated marginal means (EMMs) of capture camera share by condition, expressed as the percentage of target-capture screenshots taken from the pan-tilt camera. The bottom-camera percentage is the complement of this quantity. Error bars show 95\% confidence intervals and gray points show raw trials. Asterisks indicate Bonferroni-corrected pairwise differences (* $p<.05$, ** $p<.01$, *** $p<.001$).}
  \label{fig:capture-camera-share}
\end{figure}

Estimated marginal means indicated substantially greater pan-tilt-camera use in the head-guided camera conditions than in the joystick-camera conditions. The lowest estimated pan-tilt-camera share of target-capture screenshots occurred in \textit{JJ} ($M=81.74\%$, $SE=5.81$), followed by \textit{OJ} ($M=83.77\%$, $SE=5.29$), \textit{JH} ($M=93.34\%$, $SE=2.57$), and \textit{OH} ($M=94.17\%$, $SE=2.30$), see Fig.~\ref{fig:capture-camera-share}. Bonferroni-adjusted pairwise contrasts showed that both head-guided camera conditions differed from both joystick-camera conditions, whereas \textit{JJ} and \textit{OJ} did not differ and \textit{JH} and \textit{OH} did not differ. This indicates that when participants chose moments to capture targets, head-guided camera control led them to rely more consistently on the pan-tilt view than on the bottom fixed view. Supporting descriptives showed the same pattern in raw screenshot counts: \textit{JH} averaged $5.24$ pan-tilt-camera screenshots and $0.94$ bottom-camera screenshots, \textit{OH} averaged $5.57$ and $0.78$, while \textit{JJ} and \textit{OJ} averaged $4.88$ pan-tilt-camera screenshots paired with $1.67$ and $1.57$ bottom-camera screenshots, respectively. Sensitivity refits excluding flagged trials changed only the significance of target count and did not alter the main conclusion.

\subsubsection{Use of both cameras}
Use of both cameras was analyzed using a generalized linear mixed-effects model (GLMM) with a binomial distribution and logit link. The binary outcome indicated whether a trial included both the pan-tilt camera and the bottom fixed camera during maze traversal. Candidate additive and interaction models were compared, and the additive binomial model provided the best fit. Table~\ref{tab:supp-camera-behavior} summarizes the fixed-effect tests.

\begin{figure}[!t]
  \centering
  \includegraphics[width=\linewidth]{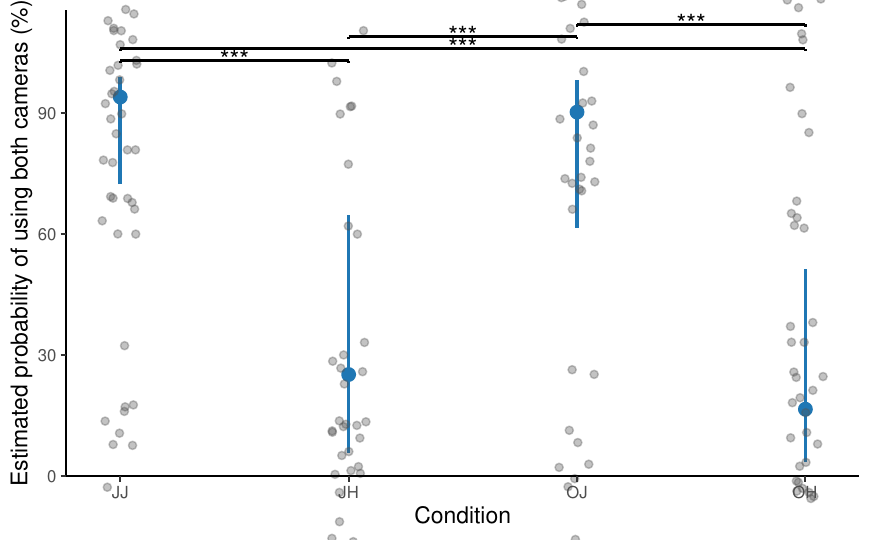}
  \caption{Estimated marginal means (EMMs) of the probability of using both cameras within a trial from the binomial GLMM (logit link). Error bars show 95\% confidence intervals and gray points show raw trials. Asterisks indicate Bonferroni-corrected pairwise differences (* $p<.05$, ** $p<.01$, *** $p<.001$).}
  \label{fig:used-both-condition}
\end{figure}

Estimated marginal means indicated that trials were much more likely to use both cameras in the joystick-camera conditions than in the head-guided-camera conditions. The highest estimated probabilities occurred in \textit{JJ} ($M=94.01\%$, $SE=5.14$) and \textit{OJ} ($M=90.28\%$, $SE=7.87$), whereas \textit{JH} ($M=25.23\%$, $SE=16.33$) and \textit{OH} ($M=16.64\%$, $SE=11.81$) were much lower, see Fig.~\ref{fig:used-both-condition}. Bonferroni-adjusted pairwise contrasts showed that both joystick-camera conditions differed from both head-guided-camera conditions (all $p\le .001$), whereas \textit{JJ} and \textit{OJ} did not differ ($p=1.000$) and \textit{JH} and \textit{OH} did not differ ($p=1.000$). This indicates that head-guided camera control substantially reduced the need to alternate between the pan-tilt and bottom views within a trial.

\begin{figure}[!t]
  \centering
  \includegraphics[width=\linewidth]{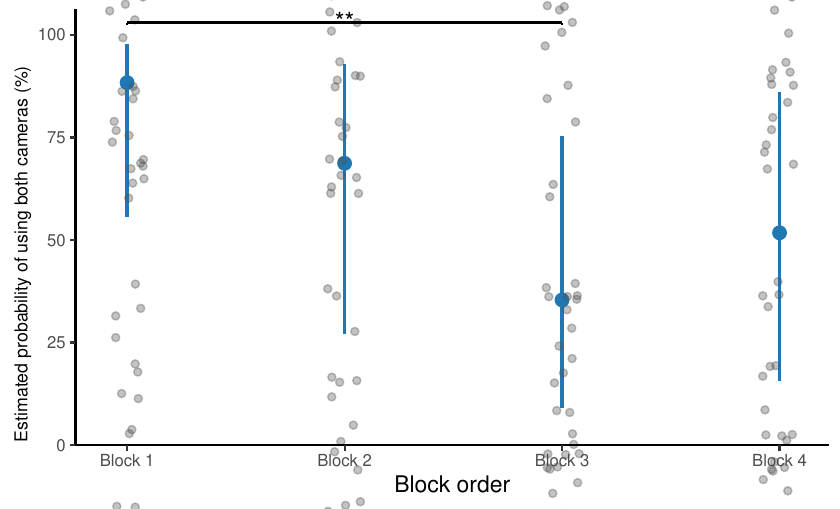}
  \caption{Estimated marginal means (EMMs) of the probability of using both cameras by block order from the binomial GLMM (logit link). Error bars show 95\% confidence intervals and gray points show raw trials. Asterisks indicate Bonferroni-corrected pairwise differences when present.}
  \label{fig:used-both-block-order}
\end{figure}

Use of both cameras also declined across the session. Estimated probabilities were highest in Block~1 ($M=88.38\%$, $SE=9.46$), then lower in Block~2 ($M=68.70\%$, $SE=19.51$), Block~4 ($M=51.77\%$, $SE=22.28$), and Block~3 ($M=35.38\%$, $SE=20.01$), indicating that participants increasingly committed to a single camera strategy rather than alternating between views. After Bonferroni correction, only the Block~1 versus Block~3 contrast remained significant ($p=.002$), see Fig.~\ref{fig:used-both-block-order}.

\begin{figure}[!t]
  \centering
  \includegraphics[width=\linewidth]{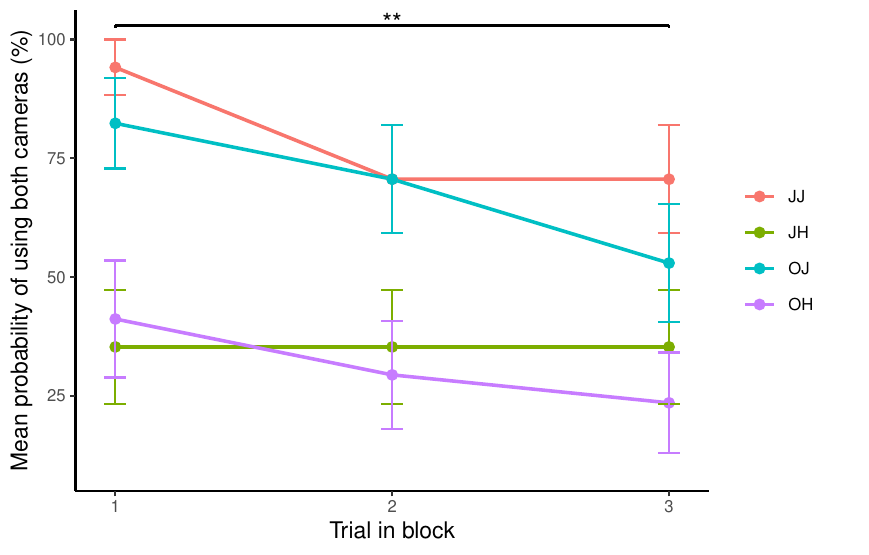}
  \caption{Estimated marginal means (EMMs) of the probability of using both cameras by trial in block from the binomial GLMM (logit link), shown separately for the four interface conditions. Error bars show 95\% confidence intervals and gray points show raw trials. Asterisks indicate Bonferroni-corrected pairwise differences when present.}
  \label{fig:used-both-learning}
\end{figure}

Within condition blocks, Trial~1 ($M=84.04\%$, $SE=11.79$) was more likely than Trial~3 ($M=40.51\%$, $SE=20.31$) to involve both cameras ($p=.006$), whereas Trial~1 versus Trial~2 and Trial~2 versus Trial~3 were not significant after correction, see Fig.~\ref{fig:used-both-learning}. Supporting descriptives on the raw trial counts showed the same pattern (JJ: $40/51$ trials; JH: $18/51$; OJ: $35/51$; OH: $16/51$). Sensitivity refits excluding flagged trials changed only the significance of target count and did not alter the main conclusion.

\subsubsection{Camera viewing time}
Absolute viewing time for each camera was analyzed separately because time spent in the pan-tilt camera and time spent in the bottom fixed camera have different distributional properties and are not simple complements once overall trial duration varies across conditions.

\begin{figure}[!t]
  \centering
  \includegraphics[width=\linewidth]{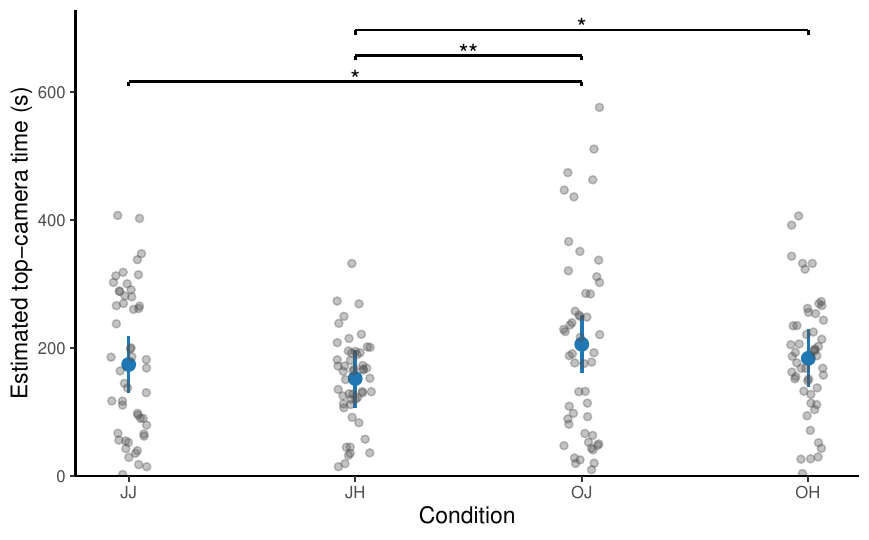}
  \caption{Estimated marginal means (EMMs) of pan-tilt-camera viewing time by condition. Error bars show 95\% confidence intervals and gray points show raw trials. Asterisks indicate Bonferroni-corrected pairwise differences (* $p<.05$, ** $p<.01$, *** $p<.001$).}
  \label{fig:pan-tilt-time-condition}
\end{figure}

Time spent viewing the pan-tilt camera was analyzed using a linear mixed-effects model (LMM) with a Gaussian distribution and identity link. Candidate Gaussian and Gamma mixed models were compared, and the additive Gaussian model provided the best fit. Table~\ref{tab:supp-camera-behavior} summarizes the fixed-effect tests. Estimated marginal means indicated that pan-tilt-camera time was lowest in \textit{JH} ($M=152.78$~s), followed by \textit{JJ} ($M=174.73$~s), \textit{OH} ($M=184.38$~s), and \textit{OJ} ($M=206.34$~s), see Fig.~\ref{fig:pan-tilt-time-condition}. Bonferroni-adjusted contrasts showed that \textit{OJ} exceeded both \textit{JJ} ($p=.010$) and \textit{JH} ($p=.001$), and that \textit{OH} exceeded \textit{JH} ($p=.010$). Absolute pan-tilt-camera time partly reflects overall completion time, so the dwell-allocation analysis gives the cleaner view of relative camera use.

\begin{figure}[!t]
  \centering
  \includegraphics[width=\linewidth]{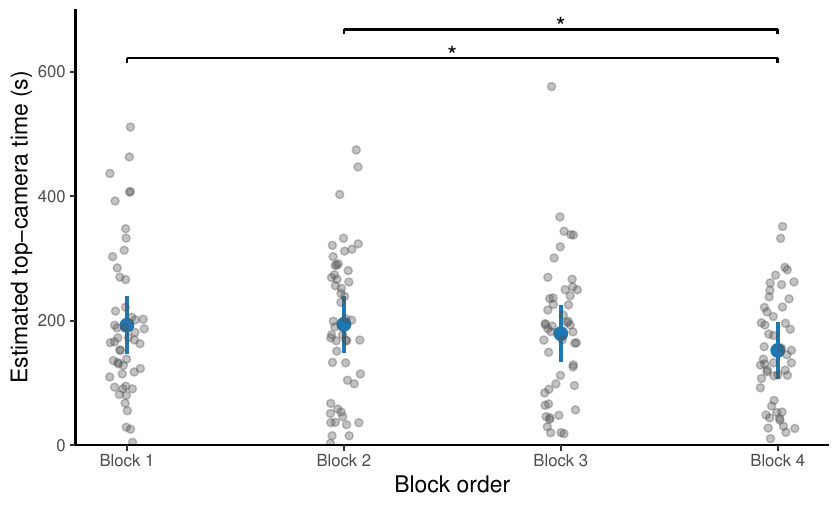}
  \caption{Estimated marginal means (EMMs) of pan-tilt-camera viewing time by block order. Error bars show 95\% confidence intervals and gray points show raw trials. Asterisks indicate Bonferroni-corrected pairwise differences when present.}
  \label{fig:pan-tilt-time-block-order}
\end{figure}

Block-order contrasts showed longer pan-tilt-camera viewing in Block~4 than in both Block~1 ($p=.025$) and Block~2 ($p=.019$), which is consistent with participants spending more of the later session actively operating through the pan-tilt view (Fig.~\ref{fig:pan-tilt-time-block-order}).

\begin{figure}[!t]
  \centering
  \includegraphics[width=\linewidth]{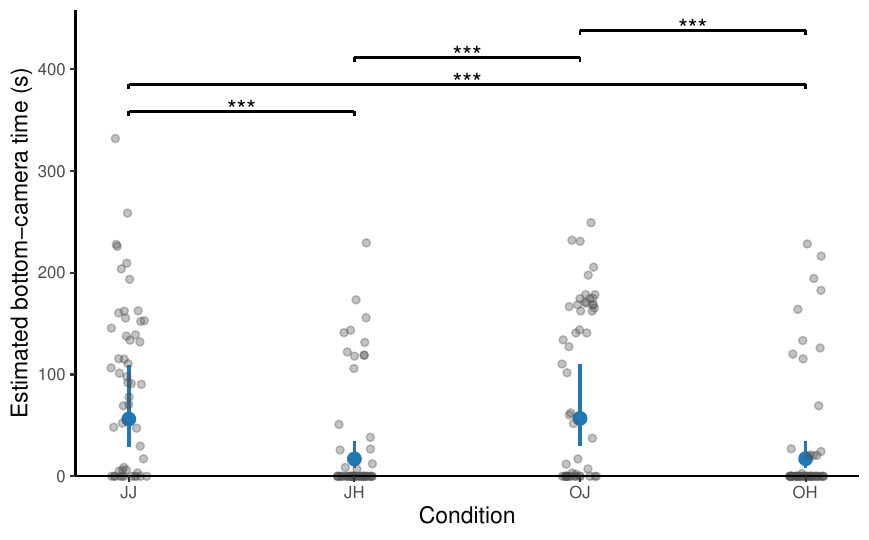}
  \caption{Estimated marginal means (EMMs) of bottom-camera viewing time by condition from the Tweedie GLMM (log link). Error bars show 95\% confidence intervals and gray points show raw trials. Asterisks indicate Bonferroni-corrected pairwise differences (* $p<.05$, ** $p<.01$, *** $p<.001$).}
  \label{fig:bottom-time-condition}
\end{figure}

Time spent viewing the bottom fixed camera was analyzed using a generalized linear mixed-effects model (GLMM) with a Tweedie distribution and log link. Candidate Gaussian and Tweedie mixed models were compared, and the additive Tweedie model provided the best fit. Table~\ref{tab:supp-camera-behavior} summarizes the fixed-effect tests. Estimated marginal means indicated substantially greater bottom-camera viewing time in the joystick-camera conditions (\textit{JJ}: $M=56.21$~s; \textit{OJ}: $M=56.87$~s) than in the head-guided camera conditions (\textit{JH}: $M=17.06$~s; \textit{OH}: $M=17.26$~s), see Fig.~\ref{fig:bottom-time-condition}. Bonferroni-adjusted contrasts showed that both joystick-camera conditions exceeded both head-guided camera conditions (all $p<.001$), whereas \textit{JJ} and \textit{OJ} did not differ and \textit{JH} and \textit{OH} did not differ. This indicates that head-guided camera control sharply reduced reliance on the bottom fixed camera.

\subsubsection{Camera Switching Across Maze Sections}
Camera switching across maze sections was analyzed using a negative-binomial generalized linear mixed-effects model (GLMM; \texttt{nbinom1}, log link). Each trial was divided into five ordered maze sections from start to end using the maze-boundary checkpoints. For interpretation, Sections~1--2 are treated as the beginning of the maze, Section~3 as the middle, and Sections~4--5 as the end. Relative to the general model structure described above, these section-level models added \textit{maze section}. Random intercepts for \textit{participant} and \textit{trial} accounted for repeated section rows within trials.

\begin{figure}[!t]
  \centering
  \includegraphics[width=\linewidth]{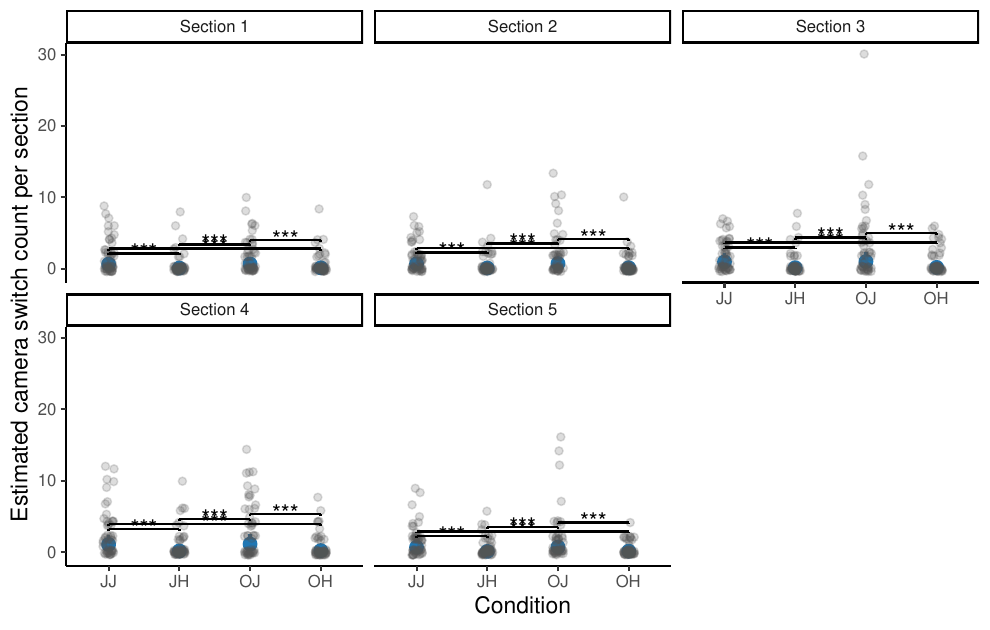}
  \caption{Estimated marginal means (EMMs) of camera switching across the five maze sections for each condition. Sections~1--2 represent the beginning of the maze, Section~3 the middle, and Sections~4--5 the end. Error bars show 95\% confidence intervals.}
  \label{fig:camera-section-switching}
\end{figure}

Table~\ref{tab:supp-camera-behavior} summarizes the fixed-effect tests and model \(R^2\) values for the section-level switching model.

Collapsing across sections, switching was much lower in the head-guided-camera conditions than in the joystick-camera conditions: JH ($M=0.13$, $SE=0.06$), OH ($M=0.17$, $SE=0.07$), JJ ($M=0.80$, $SE=0.32$), and OJ ($M=0.84$, $SE=0.34$). Bonferroni-adjusted post-hoc tests showed that both head-guided-camera conditions differed from both joystick-camera conditions, whereas JJ and OJ did not differ, and JH and OH did not differ. Switching also varied across the maze. Sections~3 and~4 showed the highest switching overall, whereas Sections~1, 2, and 5 were lower. Importantly, the condition-by-section interaction was not retained, indicating that condition differences in switching were broadly parallel across the maze. Descriptively, however, the joystick-camera conditions maintained much higher switching through the middle and end of the maze, especially OJ (Section~3 $=3.41$, Section~4 $=3.20$), whereas the head-guided-camera conditions stayed low throughout (JH range $=0.61$--$1.10$; OH range $=0.49$--$0.96$), see Fig.~\ref{fig:camera-section-switching}. This pattern suggests that joystick-camera conditions encouraged continued camera management throughout the maze, whereas head-guided-camera conditions more often allowed participants to settle on a view and maintain it. Switching also decreased with practice: section-level switch counts were highest in Block~1 and lowest in Block~3, and were higher in Trial~1 than in Trial~3 within a block. Sensitivity refits excluding flagged section rows changed only the significance of \textit{target count} and did not alter the main interpretation.

\subsection{Camera-Strategy Augmented Models}
The augmented models examine whether trial-level camera strategy helps explain the objective performance differences between interface conditions. These analyses are exploratory and provide supporting evidence for the interpretation of viewpoint-management cost. The primary mixed models remain the main basis for the results.

\subsubsection{Camera Strategy and Completion Time}
To test whether camera-use strategy accounted for part of the completion-time condition effect, we compared the primary completion-time model with an augmented model that added trial-level camera-strategy predictors. The augmented model added three exploratory predictors derived from the section-level camera data: total camera switches across the five maze sections, whether the last switch occurred in the late maze (Sections~4--5; \textit{late switch}), and mean pan-tilt-camera dwell in the late maze (Sections~4--5). Here, Sections~1--2 represent the beginning of the maze, Section~3 the middle, and Sections~4--5 the end.

\begin{figure}[!t]
  \centering
  \includegraphics[width=\linewidth]{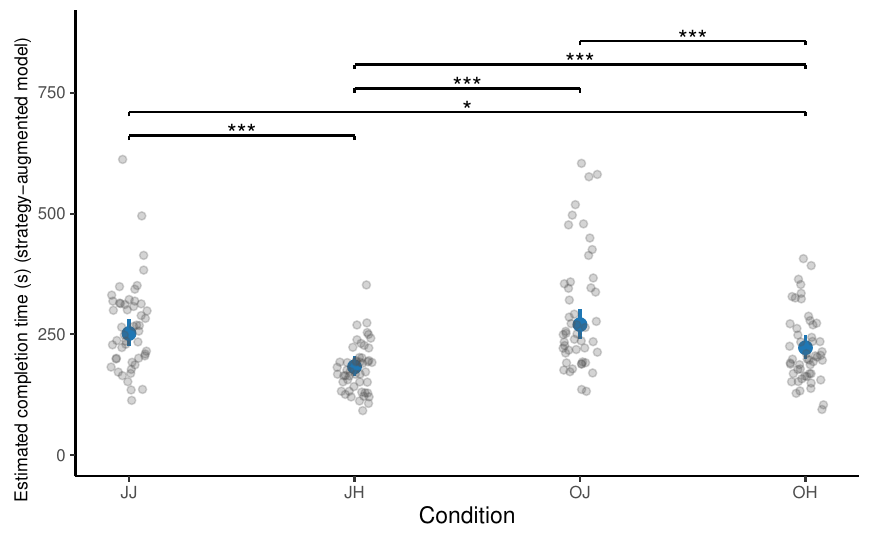}
  \caption{Adjusted condition means for completion time from the camera-strategy augmented model. The model adds total section switches, late switching, and late-maze pan-tilt-camera dwell to the baseline completion-time model. Error bars show 95\% confidence intervals.}
  \label{fig:camera-strategy-completion}
\end{figure}

In the baseline model, \textit{condition} was significant, $\chi^2(3)=138.97$, \(p<.001\), with 5 of 6 Bonferroni-adjusted pairwise condition contrasts significant. Completion time was lowest in JH ($M=177.97$~s, $SE=11.30$), followed by OH ($M=214.57$~s, $SE=13.55$), JJ ($M=256.80$~s, $SE=16.25$), and OJ ($M=283.04$~s, $SE=17.96$). After adding the camera-strategy predictors, the condition effect remained significant, $\chi^2(3)=94.93$, \(p<.001\), and 5 of 6 Bonferroni-adjusted pairwise condition contrasts remained significant. The omnibus condition $\chi^2$ decreased by $31.7\%$. In the augmented model, more camera switching predicted longer completion time (Estimate $=0.006$, \(p=.003\)), late switching predicted longer completion time (Estimate $=0.214$, \(p<.001\)), and greater late-maze pan-tilt-camera dwell also predicted longer completion time (Estimate $=0.002$, \(p=.002\)). Trials with continued camera-management demands, especially those with switching that persisted into the late maze, tended to be slower. This pattern is consistent with camera strategy explaining part of the completion-time advantage of the head-guided-camera conditions (Fig.~\ref{fig:camera-strategy-completion}). Sensitivity refits excluding flagged trial rows did not change the augmented-model significance pattern.

\subsubsection{Camera Strategy and Wall-Contact Count}
To test whether camera-use strategy accounted for part of the condition effect on wall-contact count, we compared the primary collision-count model with an augmented model that added trial-level camera-strategy predictors. The augmented model added total camera switches across the five maze sections, whether the last switch occurred in the late maze (Sections~4--5), and mean pan-tilt-camera dwell in the late maze. Sections~1--2 represent the beginning of the maze, Section~3 the middle, and Sections~4--5 the end.

\begin{figure}[!t]
  \centering
  \includegraphics[width=\linewidth]{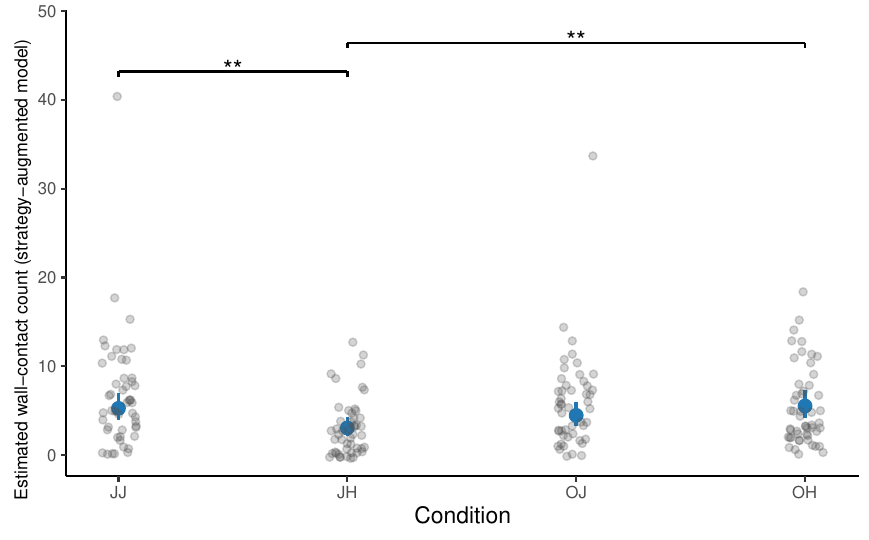}
  \caption{Adjusted condition means for wall-contact count from the camera-strategy augmented model. The model adds total section switches, late switching, and late-maze pan-tilt-camera dwell to the baseline collision-count model. Error bars show 95\% confidence intervals.}
  \label{fig:camera-strategy-collision-count}
\end{figure}

In the baseline model, \textit{condition} was significant, $\chi^2(3)=23.26$, $p<.001$, with 3 of 6 Bonferroni-adjusted pairwise condition contrasts significant. Collision count was lowest in JH ($M=2.80$, $SE=0.46$), followed by OH ($M=5.01$, $SE=0.69$), OJ ($M=5.31$, $SE=0.73$), and JJ ($M=5.99$, $SE=0.79$). After adding the camera-strategy predictors, the condition effect remained significant, $\chi^2(3)=16.53$, $p<.001$, but the omnibus condition $\chi^2$ decreased by $28.9\%$, and the number of Bonferroni-significant pairwise condition contrasts dropped from 3 to 2. More total section switches predicted more collisions (Estimate $=0.017$, $p=.025$), whereas greater late-maze pan-tilt-camera dwell predicted fewer collisions (Estimate $=-0.005$, $p=.031$). This pattern suggests that camera strategy explained part of the condition effect on collision count. More frequent switching was associated with more wall contacts, whereas greater late-maze pan-tilt-camera dwell was associated with fewer wall contacts (Fig.~\ref{fig:camera-strategy-collision-count}). Sensitivity refits excluding flagged trial rows did not change the augmented-model significance pattern.

\subsubsection{Camera Strategy and Wall-Contact Duration}
To test whether camera-use strategy accounted for part of the condition effect on wall-contact duration, we compared the primary collision-duration model with an augmented model that added trial-level camera-strategy predictors. The augmented model added total camera switches across the five maze sections, whether the last switch occurred in the late maze (Sections~4--5), and mean pan-tilt-camera dwell in the late maze. Sections~1--2 represent the beginning of the maze, Section~3 the middle, and Sections~4--5 the end.

\begin{figure}[!t]
  \centering
  \includegraphics[width=\linewidth]{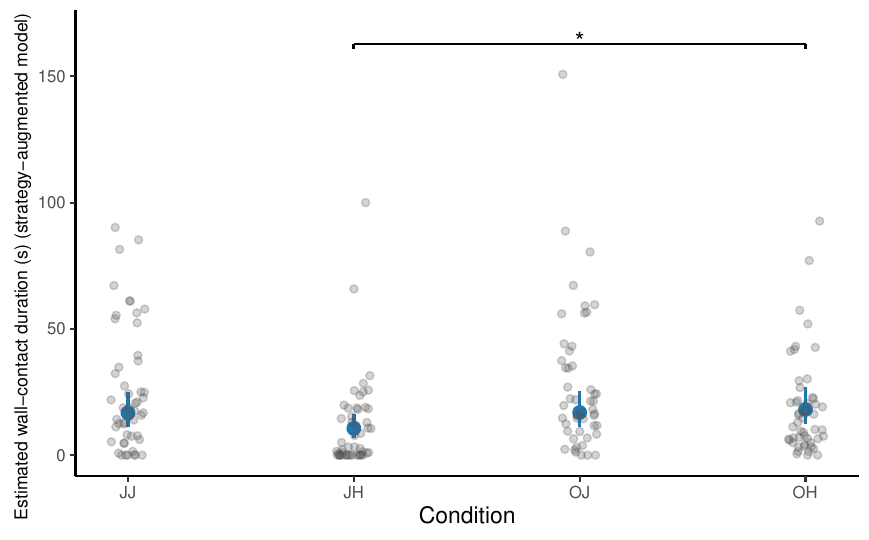}
  \caption{Adjusted condition means for wall-contact duration from the camera-strategy augmented model. The model adds total section switches, late switching, and late-maze pan-tilt-camera dwell to the baseline collision-duration model. Error bars show 95\% confidence intervals.}
  \label{fig:camera-strategy-collision-duration}
\end{figure}

In the baseline model, \textit{condition} was significant, $\chi^2(3)=22.91$, $p<.001$, with 2 of 6 Bonferroni-adjusted pairwise condition contrasts significant. Collision duration was lowest in JH ($M=9.92$~s, $SE=2.02$), followed by OH ($M=15.40$~s, $SE=2.89$), JJ ($M=20.11$~s, $SE=3.70$), and OJ ($M=23.47$~s, $SE=4.25$). After adding the camera-strategy predictors, the condition effect remained significant, but was substantially reduced, $\chi^2(3)=9.67$, $p=.022$. The omnibus condition $\chi^2$ decreased by $57.8\%$, and the number of Bonferroni-significant pairwise condition contrasts dropped from 2 to 1. Among the three performance outcomes, this model showed the strongest attenuation after adding the camera-strategy predictors. In the augmented model, more total section switches predicted longer collision duration (Estimate $=0.032$, $p<.001$). Continued camera switching was associated with more wall contacts and with longer contact duration once contacts occurred (Fig.~\ref{fig:camera-strategy-collision-duration}). Sensitivity refits excluding flagged trial rows did not change the augmented-model significance pattern.

\subsection{Validated Subjective Outcomes}
The validated questionnaire outcomes extend the main paper by reporting the full fixed-effect tests, condition estimates, and practice effects for workload, presence, simulator sickness, flow, usability, and technology acceptance. All validated subjective outcomes were analyzed with Gaussian linear mixed-effects models (LMMs) with identity link. Condition figures are shown for all measures with significant condition effects, and additional block-order or trial-in-block figures are shown where those effects were significant.

\begin{table*}[t]
\centering
\scriptsize
\setlength{\tabcolsep}{2.5pt}
\renewcommand{\arraystretch}{1.08}
\caption{Supplementary validated questionnaire model details. All outcomes used Gaussian LMMs. Effect entries report \(\chi^2(df)\), \(p\).}
\label{tab:supp-subjective}
\begin{tabular}{@{}L{0.16\textwidth}L{0.31\textwidth}L{0.35\textwidth}L{0.10\textwidth}@{}}
\toprule
Outcome & Interface effects & Task and order effects & \(R^2_m\) / \(R^2_c\) \\
\midrule
NASA-TLX &
Locomotion, \(\chi^2(1)=16.73\), \(p<.001\) \newline
Camera control, \(\chi^2(1)=32.00\), \(p<.001\) &
Maze, \(\chi^2(2)=1.31\), \(p=.519\) \newline
Target count, \(\chi^2(2)=1.64\), \(p=.440\) \newline
Block order, \(\chi^2(3)=21.46\), \(p<.001\) \newline
Trial in block, \(\chi^2(2)=7.58\), \(p=.023\) & .091 / .767 \\
\midrule
Presence (PQ) &
Locomotion, \(\chi^2(1)=23.19\), \(p<.001\) \newline
Camera control, \(\chi^2(1)=68.93\), \(p<.001\) \newline
Locomotion by camera control, \(\chi^2(1)=13.14\), \(p<.001\) &
Maze, \(\chi^2(2)=0.67\), \(p=.716\) \newline
Target count, \(\chi^2(2)=3.25\), \(p=.196\) \newline
Block order, \(\chi^2(3)=29.91\), \(p<.001\) \newline
Trial in block, \(\chi^2(2)=16.13\), \(p<.001\) & .178 / .772 \\
\midrule
Simulator sickness (SSQ) &
Locomotion, \(\chi^2(1)=0.23\), \(p=.631\) \newline
Camera control, \(\chi^2(1)=0.05\), \(p=.828\) &
Maze, \(\chi^2(2)=2.10\), \(p=.349\) \newline
Target count, \(\chi^2(2)=2.77\), \(p=.250\) \newline
Block order, \(\chi^2(3)=15.12\), \(p=.002\) \newline
Trial in block, \(\chi^2(2)=3.84\), \(p=.147\) & .022 / .826 \\
\midrule
Flow &
Locomotion, \(\chi^2(1)=7.04\), \(p=.008\) \newline
Camera control, \(\chi^2(1)=27.51\), \(p<.001\) \newline
Locomotion by camera control, \(\chi^2(1)=12.83\), \(p<.001\) &
Maze, \(\chi^2(2)=1.79\), \(p=.408\) \newline
Target count, \(\chi^2(2)=0.63\), \(p=.729\) \newline
Block order, \(\chi^2(3)=16.71\), \(p<.001\) \newline
Trial in block, \(\chi^2(2)=1.96\), \(p=.376\) & .071 / .788 \\
\midrule
System usability (SUS) &
Locomotion, \(\chi^2(1)=12.39\), \(p<.001\) \newline
Camera control, \(\chi^2(1)=78.72\), \(p<.001\) \newline
Locomotion by camera control, \(\chi^2(1)=12.57\), \(p<.001\) &
Maze, \(\chi^2(2)=0.85\), \(p=.653\) \newline
Target count, \(\chi^2(2)=0.25\), \(p=.882\) \newline
Block order, \(\chi^2(3)=8.14\), \(p=.043\) \newline
Trial in block, \(\chi^2(2)=8.85\), \(p=.012\) & .148 / .763 \\
\midrule
Technology acceptance (TAM) &
Locomotion, \(\chi^2(1)=3.96\), \(p=.047\) \newline
Camera control, \(\chi^2(1)=58.60\), \(p<.001\) \newline
Locomotion by camera control, \(\chi^2(1)=5.87\), \(p=.015\) &
Maze, \(\chi^2(2)=0.64\), \(p=.726\) \newline
Target count, \(\chi^2(2)=0.18\), \(p=.916\) \newline
Block order, \(\chi^2(3)=9.09\), \(p=.028\) \newline
Trial in block, \(\chi^2(2)=3.72\), \(p=.156\) & .124 / .706 \\
\bottomrule
\end{tabular}
\end{table*}

\subsubsection{NASA-TLX}
Table~\ref{tab:supp-subjective} summarizes the NASA-TLX fixed-effect tests and model \(R^2\) values.

\begin{figure}[!t]
  \centering
  \includegraphics[width=\linewidth]{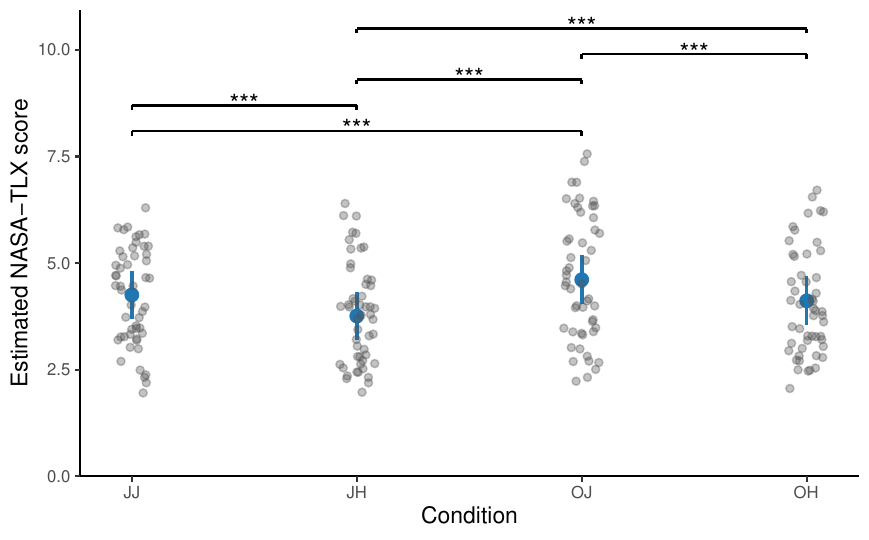}
  \caption{Estimated marginal means (EMMs) of NASA-TLX by condition from the Gaussian LMM. Error bars show 95\% confidence intervals and gray points show raw trials. Asterisks indicate Bonferroni-corrected pairwise differences (* $p<.05$, ** $p<.01$, *** $p<.001$).}
  \label{fig:nasa-tlx-condition}
\end{figure}

Workload was lowest in JH ($M=3.76$, $SE=0.27$) and highest in OJ ($M=4.61$, $SE=0.27$), with JJ ($M=4.25$, $SE=0.27$) and OH ($M=4.12$, $SE=0.27$) in between, see Fig.~\ref{fig:nasa-tlx-condition}. Bonferroni-adjusted contrasts showed that all condition pairs differed significantly except JJ versus OH. This pattern indicates that head-guided camera control reduced perceived workload, while the treadmill combined with joystick camera control produced the highest workload.

\begin{figure}[!t]
  \centering
  \includegraphics[width=\linewidth]{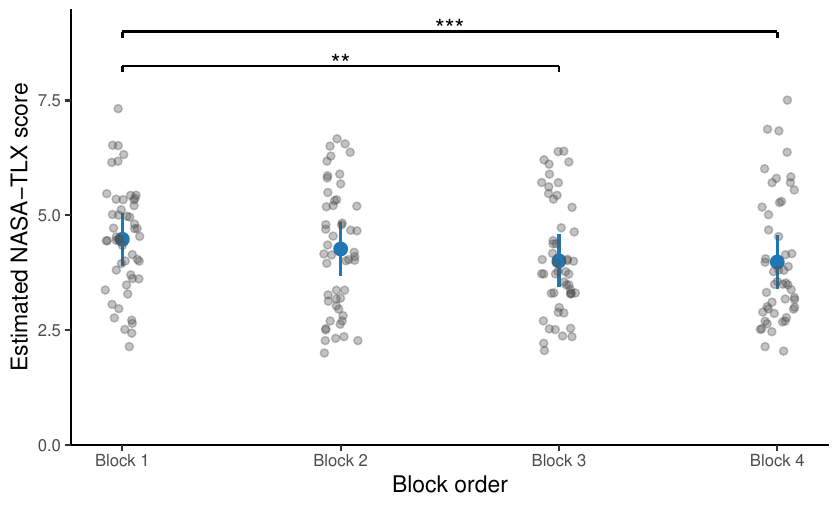}
  \caption{Estimated marginal means (EMMs) of NASA-TLX by block order from the Gaussian LMM. Error bars show 95\% confidence intervals and gray points show raw trials. Asterisks indicate Bonferroni-corrected pairwise differences when present.}
  \label{fig:nasa-tlx-block-order}
\end{figure}

Workload was also higher in Block~1 than in Blocks~3 and 4 (both $p\le .002$), indicating lower perceived effort after participants had progressed through the first part of the study (Fig.~\ref{fig:nasa-tlx-block-order}).

\begin{figure}[!t]
  \centering
  \includegraphics[width=\linewidth]{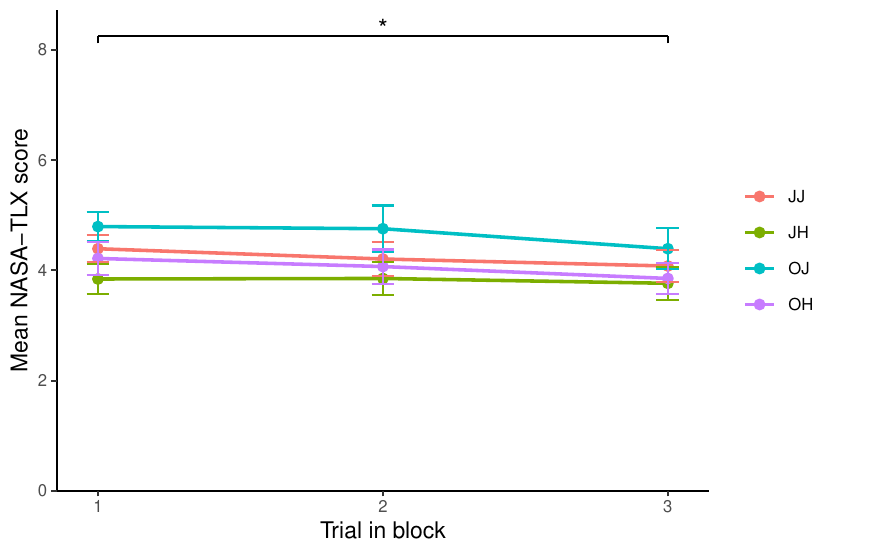}
  \caption{Estimated marginal means (EMMs) of NASA-TLX by trial in block from the Gaussian LMM, shown separately for the four interface conditions. Error bars show 95\% confidence intervals and gray points show raw trials. Asterisks indicate Bonferroni-corrected pairwise differences when present.}
  \label{fig:nasa-tlx-learning}
\end{figure}

Trial~1 was also higher than Trial~3 ($p=.029$), indicating a smaller but still detectable within-block learning effect (Fig.~\ref{fig:nasa-tlx-learning}).

\subsubsection{Presence (PQ)}
Table~\ref{tab:supp-subjective} summarizes the presence fixed-effect tests and model \(R^2\) values.

\begin{figure}[!t]
  \centering
  \includegraphics[width=\linewidth]{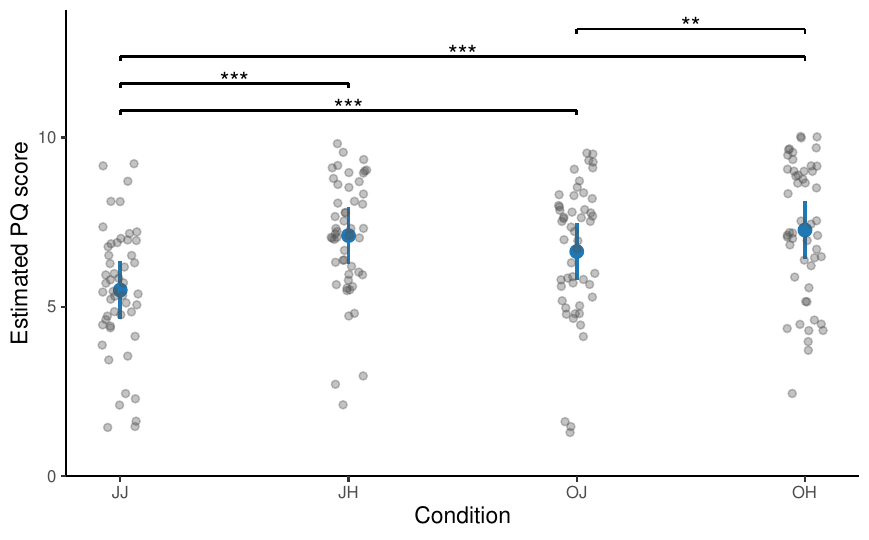}
  \caption{Estimated marginal means (EMMs) of presence (PQ) by condition from the Gaussian LMM. Error bars show 95\% confidence intervals and gray points show raw trials. Asterisks indicate Bonferroni-corrected pairwise differences (* $p<.05$, ** $p<.01$, *** $p<.001$).}
  \label{fig:pq-condition}
\end{figure}

Presence was lowest in JJ ($M=5.50$, $SE=0.41$), followed by OJ ($M=6.63$, $SE=0.41$), JH ($M=7.10$, $SE=0.41$), and OH ($M=7.27$, $SE=0.41$), see Fig.~\ref{fig:pq-condition}. Bonferroni-adjusted contrasts showed that JJ was lower than OJ ($p<.001$), JH ($p<.001$), and OH ($p<.001$), and that OJ was lower than OH ($p=.009$). The condition means indicate a monotonic improvement from the fully joystick-controlled setup to the head-guided conditions.

\begin{figure}[!t]
  \centering
  \includegraphics[width=\linewidth]{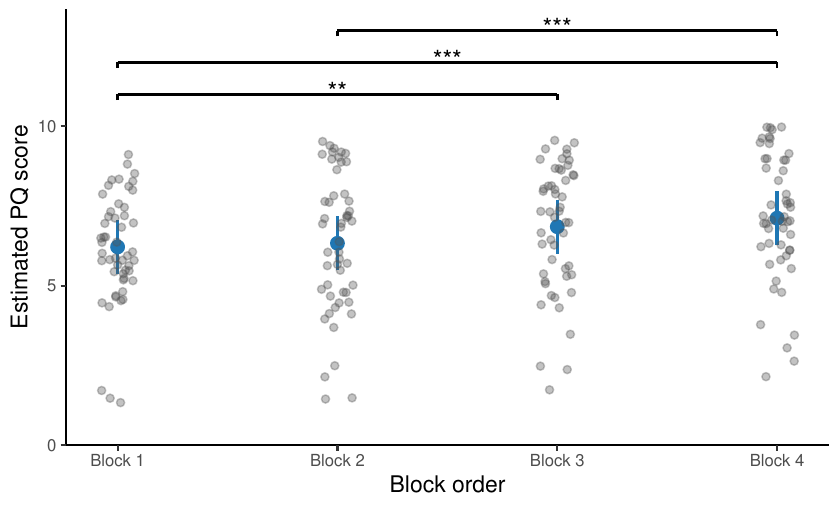}
  \caption{Estimated marginal means (EMMs) of presence (PQ) by block order from the Gaussian LMM. Error bars show 95\% confidence intervals and gray points show raw trials. Asterisks indicate Bonferroni-corrected pairwise differences when present.}
  \label{fig:pq-block-order}
\end{figure}

PQ also increased from Block~1 to Block~3 ($p=.009$) and Block~4 ($p<.001$), and from Block~2 to Block~4 ($p=.001$), indicating that presence strengthened as participants became more accustomed to the task and interface (Fig.~\ref{fig:pq-block-order}).

\begin{figure}[!t]
  \centering
  \includegraphics[width=\linewidth]{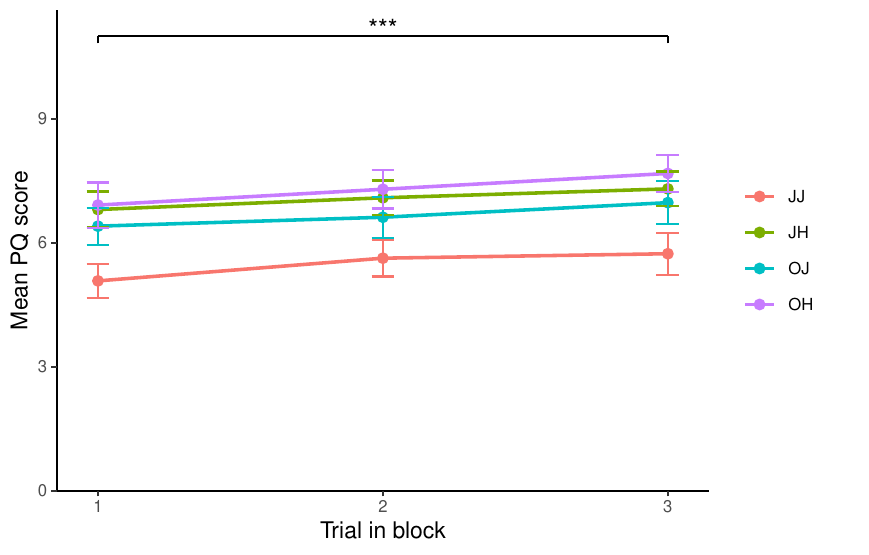}
  \caption{Estimated marginal means (EMMs) of presence (PQ) by trial in block from the Gaussian LMM, shown separately for the four interface conditions. Error bars show 95\% confidence intervals and gray points show raw trials. Asterisks indicate Bonferroni-corrected pairwise differences when present.}
  \label{fig:pq-learning}
\end{figure}

Trial~1 was lower than Trial~3 ($p<.001$), indicating that the subjective sense of presence also improved within a condition block (Fig.~\ref{fig:pq-learning}).

\subsubsection{Simulator sickness (SSQ)}
Table~\ref{tab:supp-subjective} summarizes the SSQ fixed-effect tests and model \(R^2\) values. Condition means remained low and similar across interfaces (JH $M=0.15$, JJ $M=0.16$, OH $M=0.16$, OJ $M=0.17$), and no condition contrast was significant after Bonferroni correction.

\begin{figure}[!t]
  \centering
  \includegraphics[width=\linewidth]{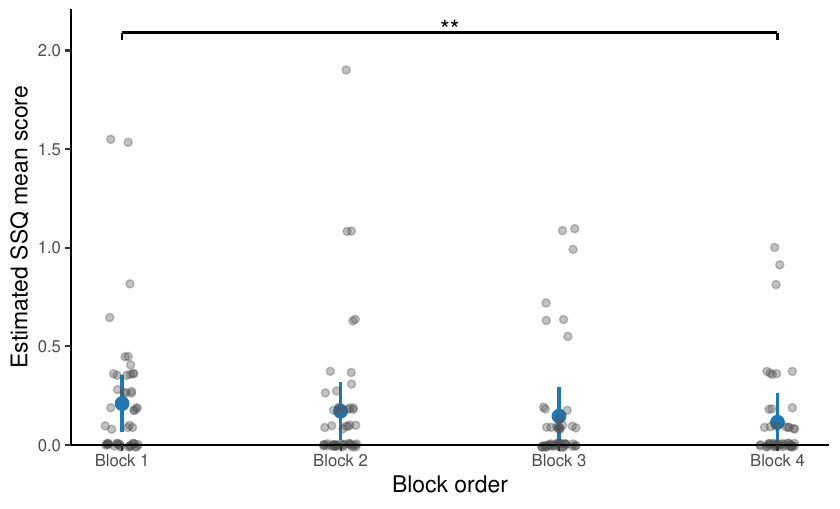}
  \caption{Estimated marginal means (EMMs) of simulator sickness (SSQ) by block order from the Gaussian LMM. Error bars show 95\% confidence intervals and gray points show raw trials. Asterisks indicate Bonferroni-corrected pairwise differences when present.}
  \label{fig:ssq-block-order}
\end{figure}

The only Bonferroni-significant block-order contrast was Block~1 versus Block~4 ($p=.002$), indicating lower SSQ scores later in the session and suggesting adaptation rather than increasing discomfort over time (Fig.~\ref{fig:ssq-block-order}).

\subsubsection{Flow State Scale (FSS)}
Table~\ref{tab:supp-subjective} summarizes the flow fixed-effect tests and model \(R^2\) values.

\begin{figure}[!t]
  \centering
  \includegraphics[width=\linewidth]{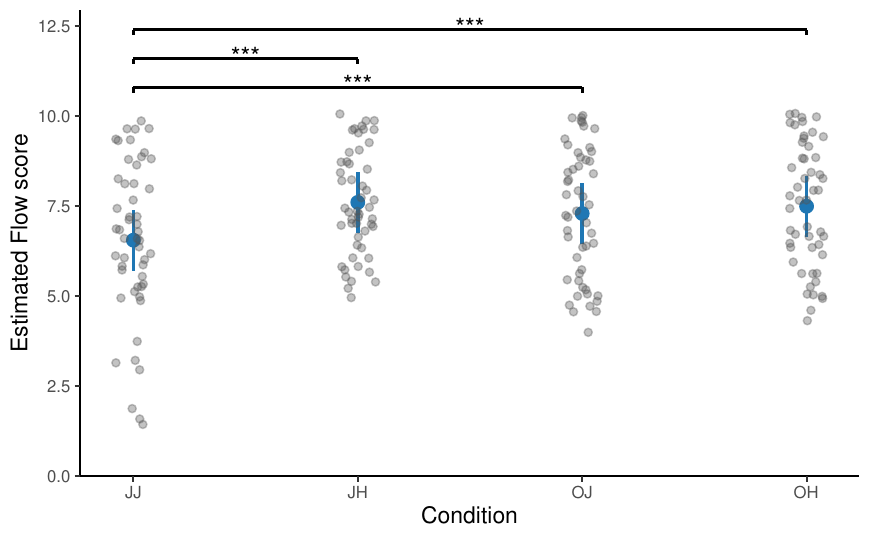}
  \caption{Estimated marginal means (EMMs) of flow by condition from the Gaussian LMM. Error bars show 95\% confidence intervals and gray points show raw trials. Asterisks indicate Bonferroni-corrected pairwise differences (* $p<.05$, ** $p<.01$, *** $p<.001$).}
  \label{fig:flow-condition}
\end{figure}

Flow was lowest in JJ ($M=6.56$, $SE=0.41$), and higher in OJ ($M=7.30$, $SE=0.41$), OH ($M=7.50$, $SE=0.41$), and JH ($M=7.61$, $SE=0.41$), see Fig.~\ref{fig:flow-condition}. Bonferroni-adjusted contrasts showed that JJ was lower than OJ ($p<.001$), OH ($p<.001$), and JH ($p<.001$), whereas the remaining condition contrasts were not significant. The means indicate lower fluent control in JJ than in the other three interfaces.

\begin{figure}[!t]
  \centering
  \includegraphics[width=\linewidth]{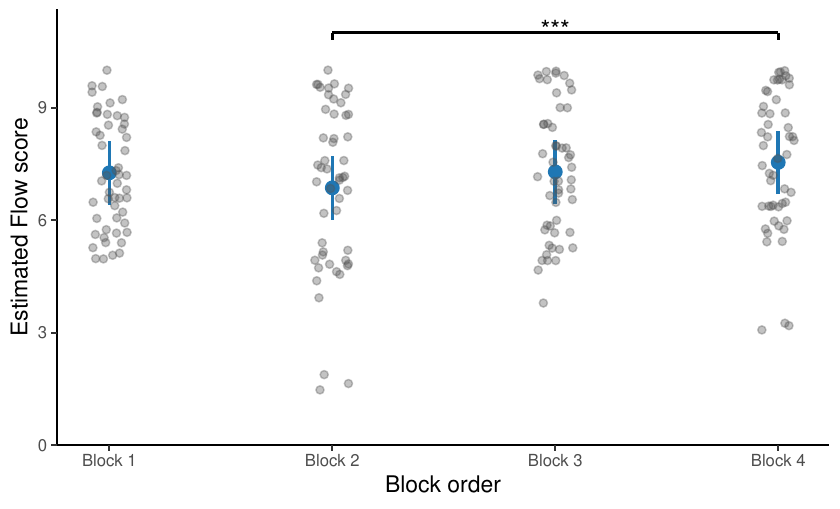}
  \caption{Estimated marginal means (EMMs) of flow by block order from the Gaussian LMM. Error bars show 95\% confidence intervals and gray points show raw trials. Asterisks indicate Bonferroni-corrected pairwise differences when present.}
  \label{fig:flow-block-order}
\end{figure}

For block order, only Block~2 versus Block~4 remained significant after correction ($p<.001$), indicating that the subjective experience of flow strengthened later in the session (Fig.~\ref{fig:flow-block-order}).

\subsubsection{System usability (SUS)}
Table~\ref{tab:supp-subjective} summarizes the SUS fixed-effect tests and model \(R^2\) values.

\begin{figure}[!t]
  \centering
  \includegraphics[width=\linewidth]{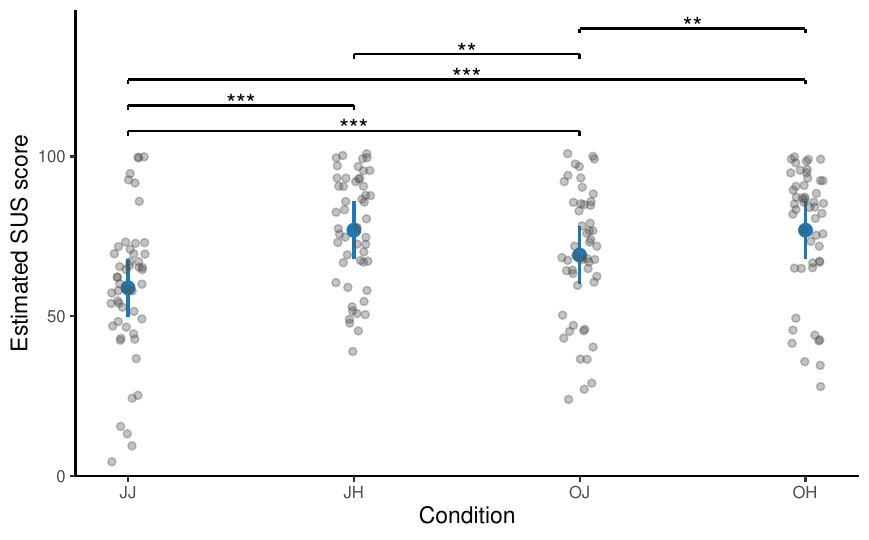}
  \caption{Estimated marginal means (EMMs) of system usability (SUS) by condition from the Gaussian LMM. Error bars show 95\% confidence intervals and gray points show raw trials. Asterisks indicate Bonferroni-corrected pairwise differences (* $p<.05$, ** $p<.01$, *** $p<.001$).}
  \label{fig:sus-condition}
\end{figure}

Usability was lowest in JJ ($M=58.93$, $SE=4.41$), intermediate in OJ ($M=69.17$, $SE=4.41$), and highest in both head-guided conditions (JH $M=76.96$, $SE=4.41$; OH $M=76.95$, $SE=4.41$), see Fig.~\ref{fig:sus-condition}. Bonferroni-adjusted contrasts showed that JJ was lower than OJ ($p<.001$), JH ($p<.001$), and OH ($p<.001$), and OJ was lower than JH ($p=.002$) and OH ($p=.002$); JH and OH did not differ. The usability means were higher in the head-guided camera conditions across both locomotion modes.

\begin{figure}[!t]
  \centering
  \includegraphics[width=\linewidth]{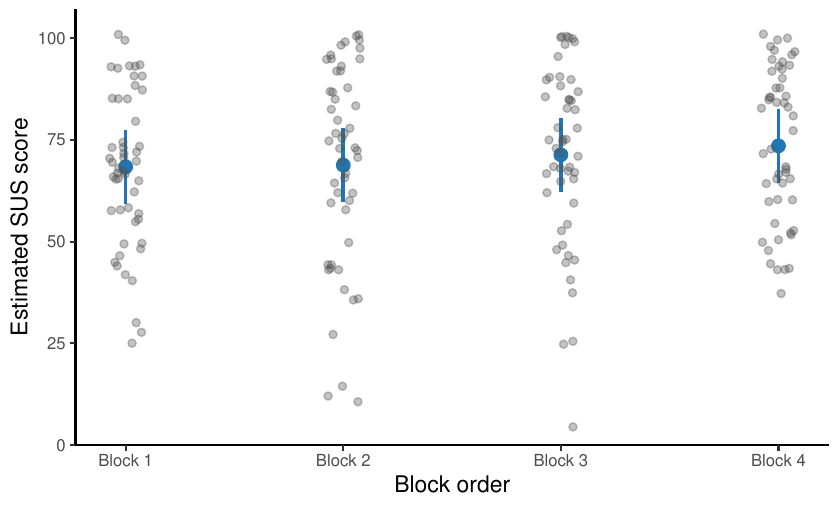}
  \caption{Estimated marginal means (EMMs) of system usability (SUS) by block order from the Gaussian LMM. Error bars show 95\% confidence intervals and gray points show raw trials. Asterisks indicate Bonferroni-corrected pairwise differences when present.}
  \label{fig:sus-block-order}
\end{figure}

Although the omnibus block-order effect was significant, no Bonferroni-adjusted pairwise block-order contrast remained significant, indicating a weak and diffuse order trend rather than a stable step change (Fig.~\ref{fig:sus-block-order}).

\begin{figure}[!t]
  \centering
  \includegraphics[width=\linewidth]{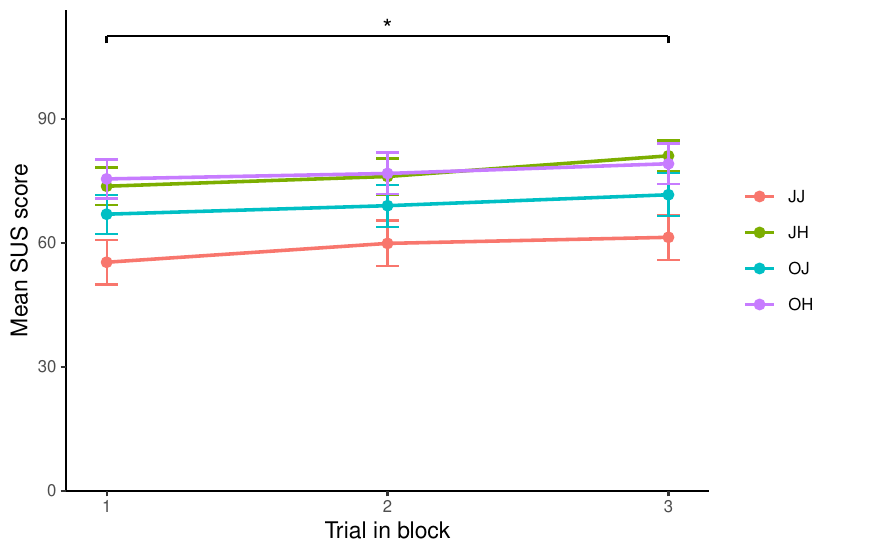}
  \caption{Estimated marginal means (EMMs) of system usability (SUS) by trial in block from the Gaussian LMM, shown separately for the four interface conditions. Error bars show 95\% confidence intervals and gray points show raw trials. Asterisks indicate Bonferroni-corrected pairwise differences when present.}
  \label{fig:sus-learning}
\end{figure}

The omnibus trial-in-block effect did not remain robust in sensitivity analysis, so interpretation focuses on the stable condition effects rather than within-block usability change (Fig.~\ref{fig:sus-learning}).

\subsubsection{Technology acceptance (TAM)}
Table~\ref{tab:supp-subjective} summarizes the TAM fixed-effect tests and model \(R^2\) values.

\begin{figure}[!t]
  \centering
  \includegraphics[width=\linewidth]{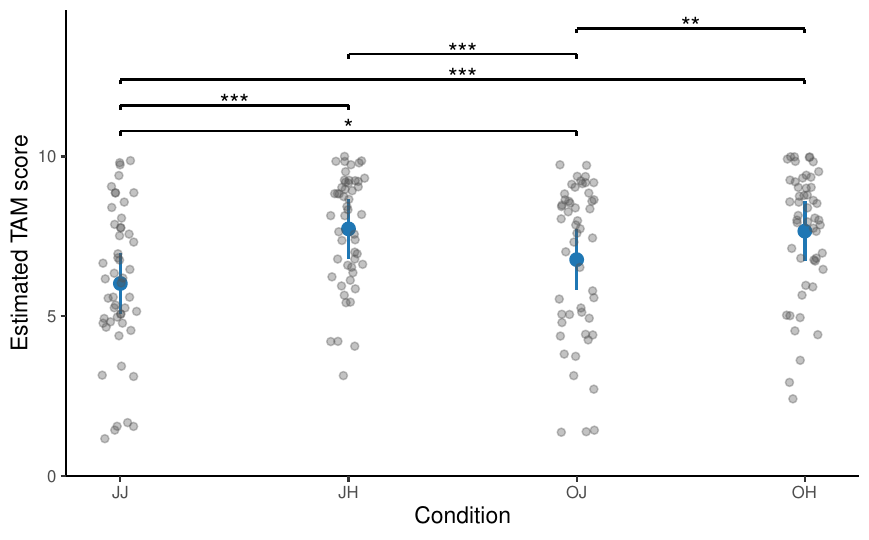}
  \caption{Estimated marginal means (EMMs) of technology acceptance (TAM) by condition from the Gaussian LMM. Error bars show 95\% confidence intervals and gray points show raw trials. Asterisks indicate Bonferroni-corrected pairwise differences (* $p<.05$, ** $p<.01$, *** $p<.001$).}
  \label{fig:tam-condition}
\end{figure}

TAM was lowest in JJ ($M=6.03$, $SE=0.46$), followed by OJ ($M=6.78$, $SE=0.46$), OH ($M=7.67$, $SE=0.46$), and JH ($M=7.74$, $SE=0.46$), see Fig.~\ref{fig:tam-condition}. Bonferroni-adjusted contrasts showed that JJ was lower than OJ ($p=.017$), JH ($p<.001$), and OH ($p<.001$), and OJ was lower than JH ($p<.001$) and OH ($p=.002$). JH and OH did not differ. Technology acceptance was therefore higher for the two head-guided camera conditions than for the joystick-camera conditions.

\begin{figure}[!t]
  \centering
  \includegraphics[width=\linewidth]{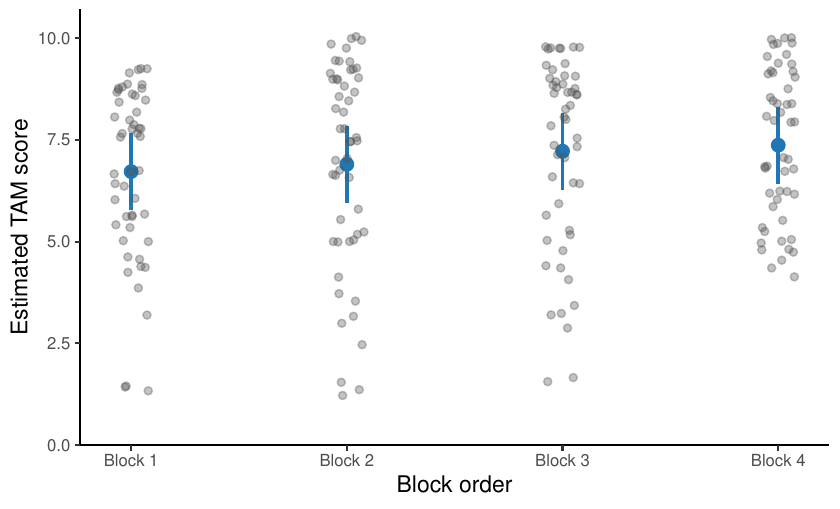}
  \caption{Estimated marginal means (EMMs) of technology acceptance (TAM) by block order from the Gaussian LMM. Error bars show 95\% confidence intervals and gray points show raw trials. Asterisks indicate Bonferroni-corrected pairwise differences when present.}
  \label{fig:tam-block-order}
\end{figure}

Although the omnibus block-order effect was significant, no Bonferroni-adjusted pairwise block-order contrast remained significant, indicating a weaker order pattern than the condition effect itself (Fig.~\ref{fig:tam-block-order}).

\subsection{Exploratory Custom and Post-Study Responses}
The custom and post-study questionnaires were exploratory. They are reported here because they help interpret how participants described coordination, camera reliance, difficulty, preference, and fatigue after using the four interfaces.

\subsubsection{Custom post-trial items}
\begin{figure}[!t]
  \centering
  \includegraphics[width=\linewidth]{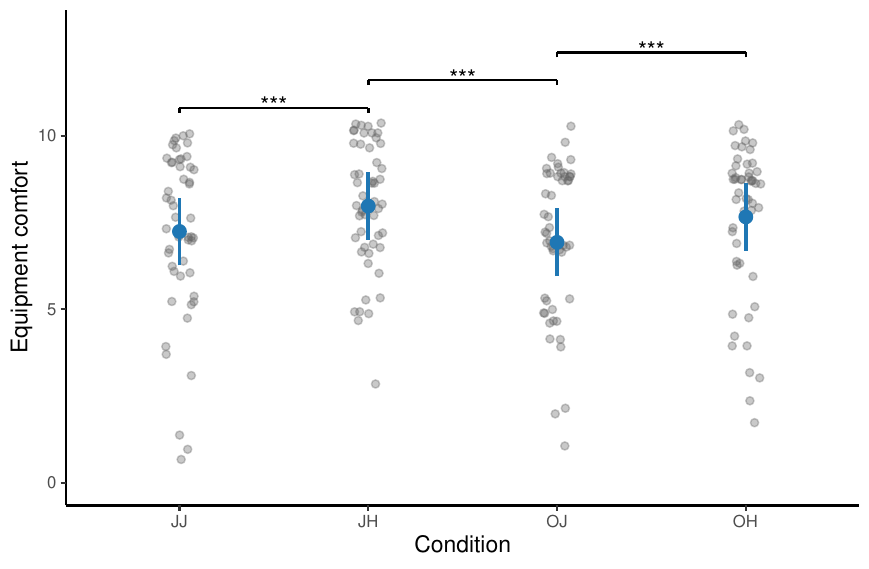}
  \caption{Exploratory post-trial equipment comfort ratings by condition. Higher values indicate greater comfort. Error bars show uncertainty around the condition summaries.}
  \label{fig:task-specific-comfort}
\end{figure}

\begin{figure}[!t]
  \centering
  \includegraphics[width=\linewidth]{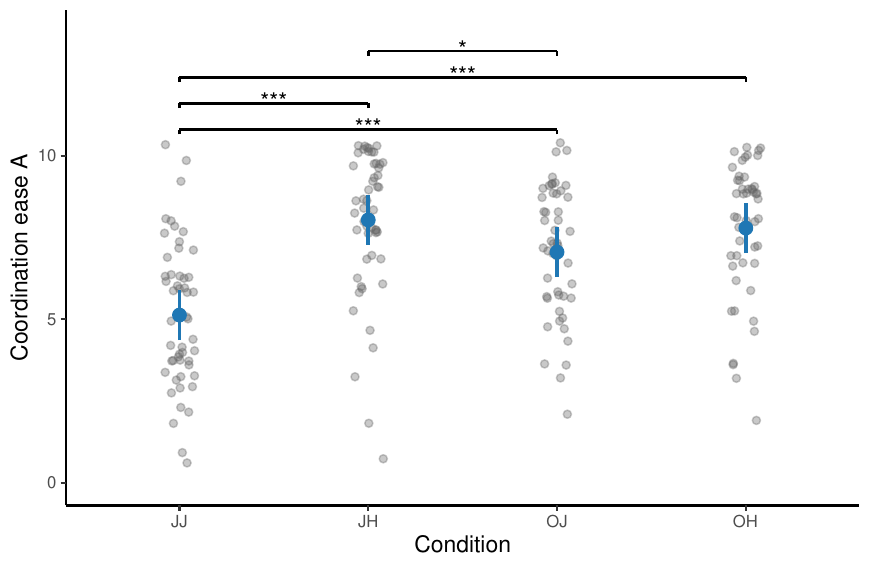}
  \caption{Exploratory post-trial coordination ease ratings for item A by condition. Higher values indicate easier perceived coordination. Error bars show uncertainty around the condition summaries.}
  \label{fig:task-specific-coord-a}
\end{figure}

\begin{figure}[!t]
  \centering
  \includegraphics[width=\linewidth]{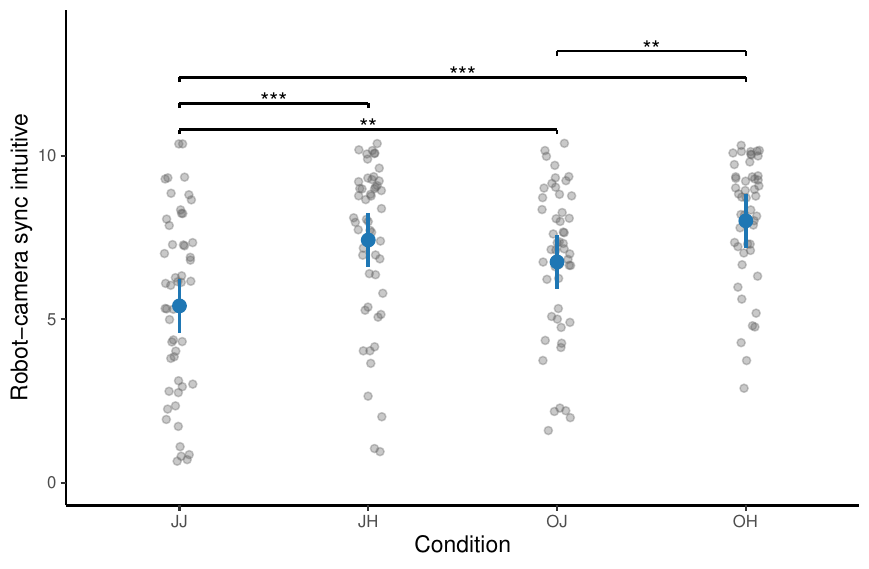}
  \caption{Exploratory post-trial robot--camera synchronization intuitiveness ratings by condition. Higher values indicate more intuitive synchronization. Error bars show uncertainty around the condition summaries.}
  \label{fig:task-specific-sync-intuitive}
\end{figure}

\begin{figure}[!t]
  \centering
  \includegraphics[width=\linewidth]{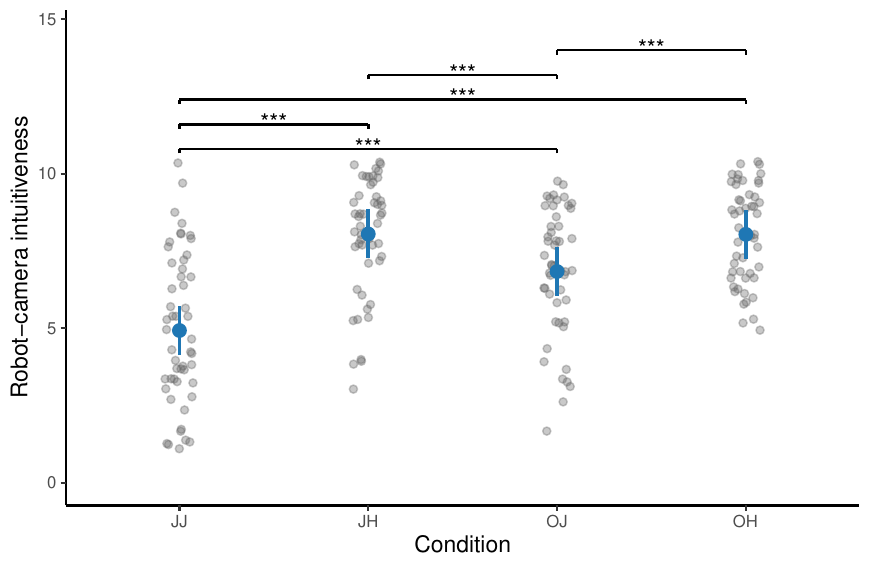}
  \caption{Exploratory post-trial robot--camera intuitiveness ratings by condition. Higher values indicate greater intuitive control. Error bars show uncertainty around the condition summaries.}
  \label{fig:task-specific-intuitive}
\end{figure}

\begin{figure}[!t]
  \centering
  \includegraphics[width=\linewidth]{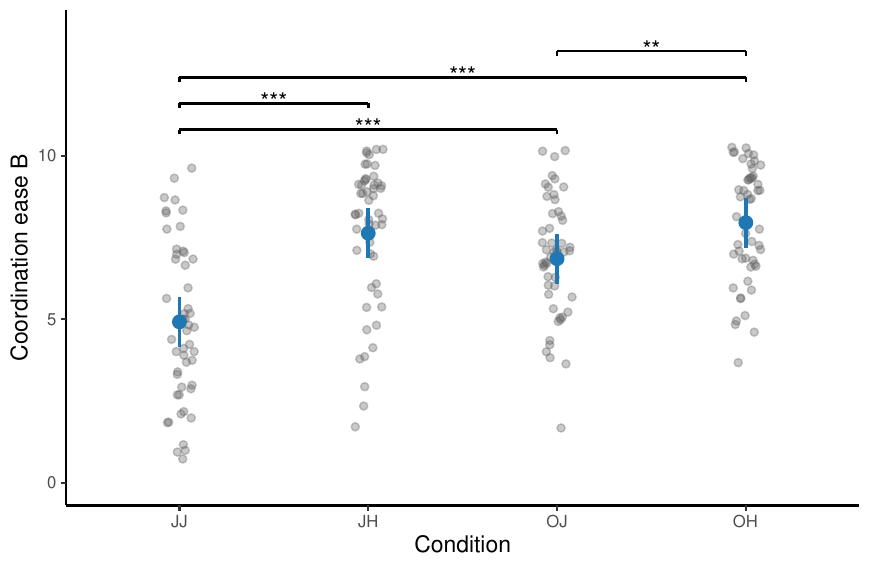}
  \caption{Exploratory post-trial coordination ease ratings for item B by condition. Higher values indicate easier perceived coordination. Error bars show uncertainty around the condition summaries.}
  \label{fig:task-specific-coord-b}
\end{figure}

\begin{figure}[!t]
  \centering
  \includegraphics[width=\linewidth]{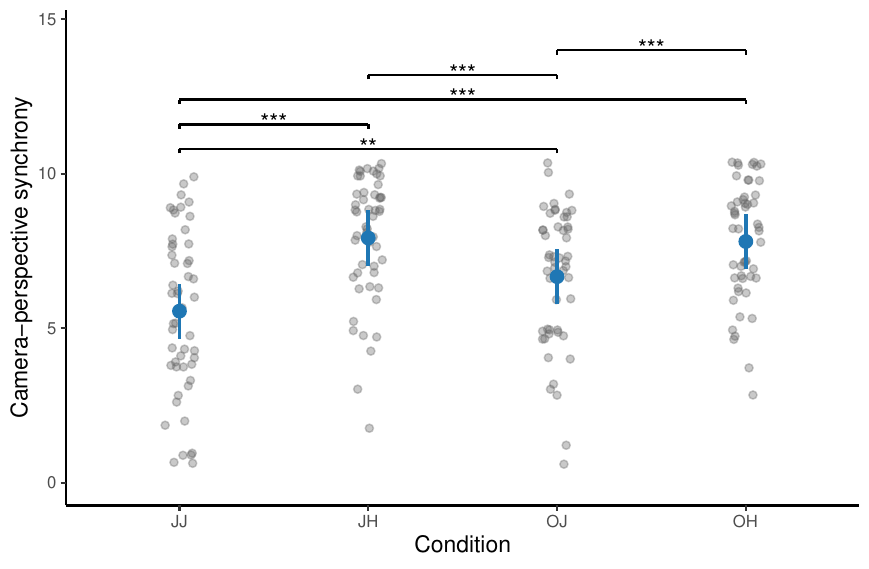}
  \caption{Exploratory post-trial camera-perspective synchrony ratings by condition. Higher values indicate stronger perceived synchrony between viewpoint and control. Error bars show uncertainty around the condition summaries.}
  \label{fig:task-specific-camera-sync}
\end{figure}

The custom post-trial items were treated as exploratory. JJ received the lowest ratings on coordination, intuitiveness, and synchronization items, whereas the two head-guided camera conditions received the highest ratings. Robot--camera intuitiveness was lowest in JJ ($M=4.93$, $SE=0.39$) and highest in JH and OH (both $M\approx8.05$). Coordination ease showed a similar progression from JJ ($M=5.13$, $SE=0.37$) to OJ ($M=7.06$, $SE=0.37$), JH ($M=8.04$, $SE=0.38$), and OH ($M=7.79$, $SE=0.37$). Equipment comfort also favored the head-guided conditions, although the difference was smaller (OJ: $M=6.93$, $SE=0.47$; JH: $M=7.97$, $SE=0.47$). These exploratory numeric items were consistent with the validated scales in showing higher perceived coordination and intuitiveness for head-guided camera control (Figs.~\ref{fig:task-specific-comfort}, \ref{fig:task-specific-coord-a}, \ref{fig:task-specific-sync-intuitive}, \ref{fig:task-specific-intuitive}, \ref{fig:task-specific-coord-b}, and~\ref{fig:task-specific-camera-sync}).

\begin{figure}[!t]
  \centering
  \includegraphics[width=\linewidth]{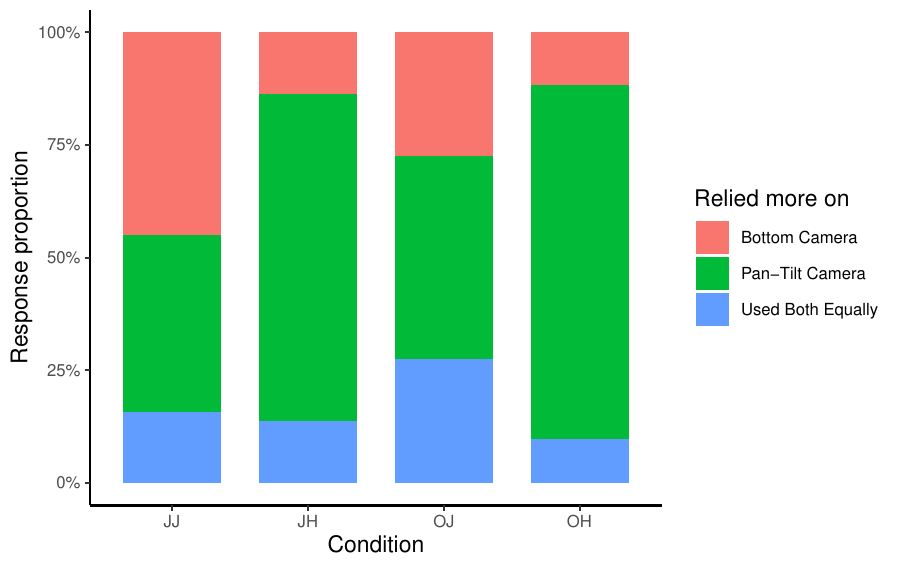}
  \caption{Exploratory distribution of reported camera reliance by condition. Bars show response counts for the reported primary view.}
  \label{fig:task-specific-camera-reliance}
\end{figure}

\begin{figure}[!t]
  \centering
  \includegraphics[width=\linewidth]{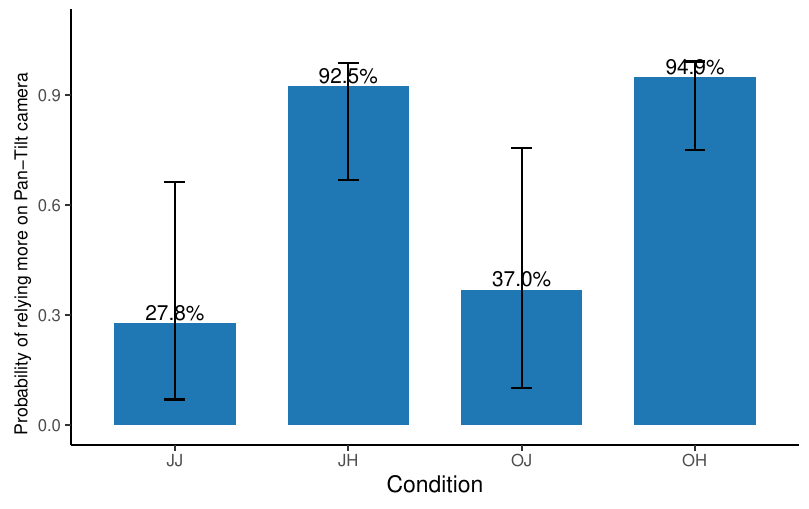}
  \caption{Exploratory estimated probability of relying primarily on the pan-tilt camera by condition from the binomial GLMM. Error bars show uncertainty around the condition estimates.}
  \label{fig:task-specific-pan-reliance}
\end{figure}

The categorical custom responses also showed differences in which view participants reported relying on while navigating. Descriptively, Pan-Tilt Camera responses dominated in both head-guided camera conditions (JH: 37/51; OH: 40/51), whereas joystick-camera conditions showed a more mixed distribution (JJ: 20/51; OJ: 23/51). In an exploratory binomial GLMM collapsing responses to Pan-Tilt Camera versus other responses, camera control was significant, $\chi^2(1)=24.17$, \(p<.001\). Estimated probabilities were 0.28 for JJ, 0.37 for OJ, 0.92 for JH, and 0.95 for OH, with Bonferroni-adjusted contrasts showing that both head-guided conditions differed from both joystick-camera conditions (Figs.~\ref{fig:task-specific-camera-reliance} and~\ref{fig:task-specific-pan-reliance}).

\subsubsection{Post-study questionnaire}
\begin{figure}[!t]
  \centering
  \includegraphics[width=\linewidth]{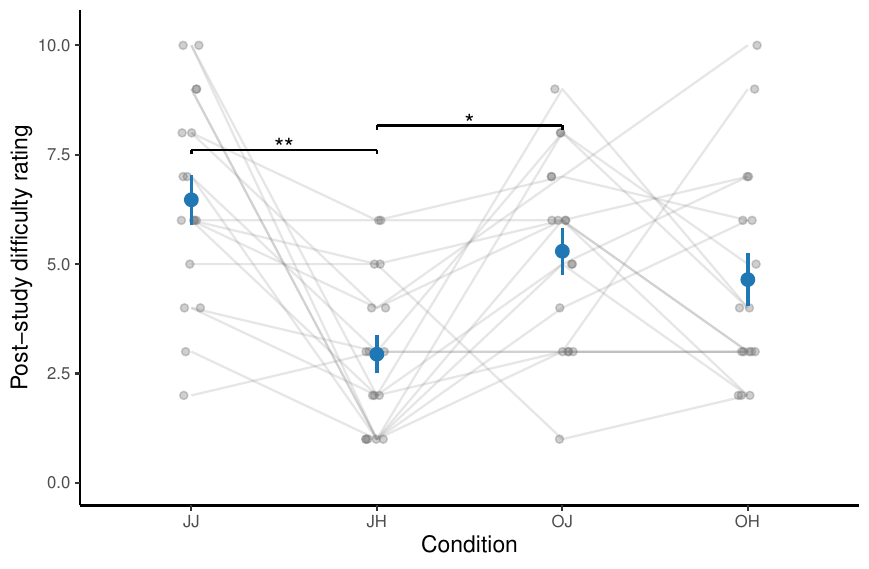}
  \caption{Post-study difficulty ratings by condition. Lower values indicate easier overall use. Error bars show uncertainty around the condition summaries.}
  \label{fig:post-study-difficulty}
\end{figure}

Participants completed a post-study questionnaire after finishing all four interface conditions. Post-study difficulty differed across the four interfaces, Friedman $\chi^2(3)=16.73$, \(p<.001\). Mean difficulty ratings were highest for JJ ($M=6.47$), followed by OJ ($M=5.29$), OH ($M=4.65$), and JH ($M=2.94$), indicating that JH was judged easiest overall. Bonferroni-adjusted Wilcoxon signed-rank tests showed that JH was rated easier than JJ ($p=.006$) and OJ ($p=.045$), whereas the remaining pairwise contrasts were not significant (Fig.~\ref{fig:post-study-difficulty}).

\begin{figure}[!t]
  \centering
  \includegraphics[width=\linewidth]{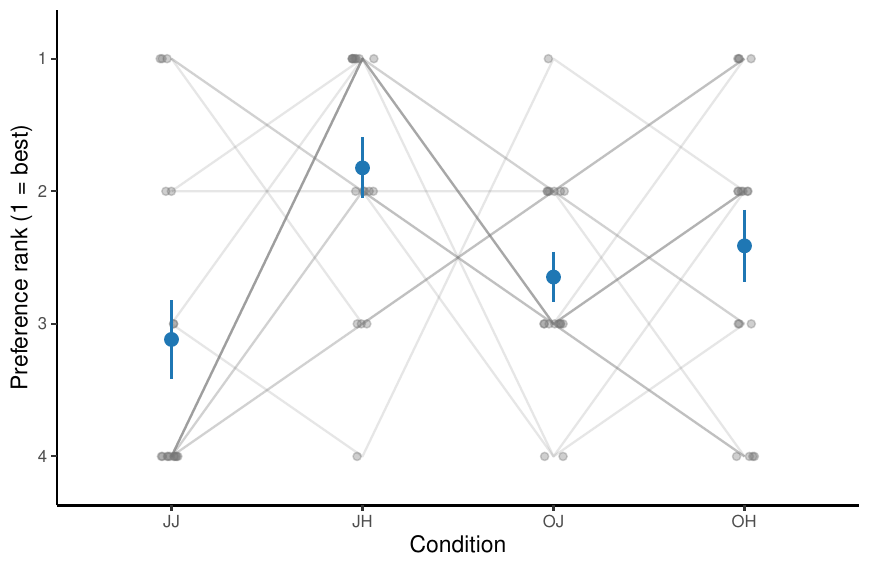}
  \caption{Post-study preference rankings by condition. Lower ranks indicate stronger overall preference. Error bars show uncertainty around the condition summaries.}
  \label{fig:post-study-preference}
\end{figure}

\begin{figure}[!t]
  \centering
  \includegraphics[width=\linewidth]{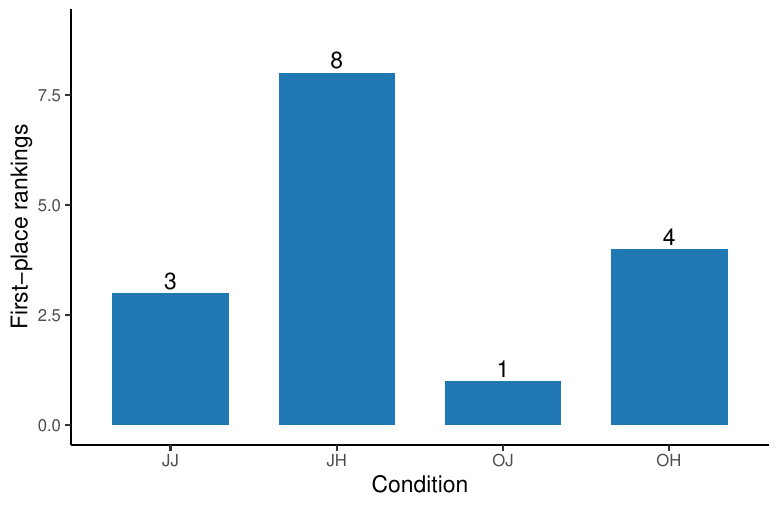}
  \caption{First-place preference rankings assigned to each condition in the post-study questionnaire. Bars show participant counts.}
  \label{fig:post-study-first-place}
\end{figure}

\begin{figure}[!t]
  \centering
  \includegraphics[width=\linewidth]{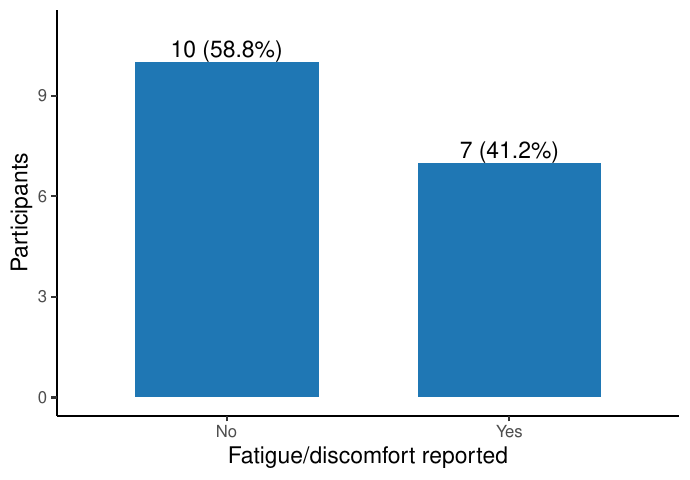}
  \caption{Post-study fatigue or discomfort responses. Bars show the number of participants reporting fatigue or discomfort versus no such report.}
  \label{fig:post-study-fatigue}
\end{figure}

Preference rankings also differed overall, Friedman $\chi^2(3)=9.58$, \(p=.022\), where lower ranks indicate stronger preference. Descriptively, JH had the best mean rank ($M=1.82$) and received the most first-place rankings (8 of 17 participants), whereas JJ had the weakest mean rank ($M=3.12$) and was ranked fourth by 10 participants. No Bonferroni-adjusted pairwise preference contrast was significant, so this pattern is reported descriptively. Finally, 7 of 17 participants (41.2\%) reported some fatigue or discomfort in the post-study questionnaire. These responses are treated as exploratory qualitative feedback because they were not formally coded by condition (Figs.~\ref{fig:post-study-preference}, \ref{fig:post-study-first-place}, and~\ref{fig:post-study-fatigue}).

